\pdfoutput=1
\documentclass{article} 

    \PassOptionsToPackage{numbers, compress}{natbib}

\usepackage[preprint]{neurips_2024}

\usepackage[utf8]{inputenc} %
\usepackage[T1]{fontenc}    %
\usepackage{hyperref}       %
\usepackage{url}            %
\usepackage{booktabs}       %
\usepackage{amsfonts}       %
\usepackage{nicefrac}       %
\usepackage{microtype}      %
\usepackage{xcolor}         %

\newcounter{packednmbr}

\usepackage{bm}
\usepackage{hhline}
\usepackage{colortbl}
\usepackage{subcaption}
\usepackage{wrapfig}
\usepackage{titletoc}
\usepackage{graphicx}
\usepackage{enumitem}
\usepackage{verbatim}
\usepackage{mdframed}
\usepackage[most]{tcolorbox}
\usepackage{listings}
\usepackage{color}
\usepackage{xcolor}
\usepackage{xspace}
\usepackage{bigstrut}

\usepackage{algorithm}
\usepackage{algorithmic}
\usepackage{amssymb}
\usepackage{graphicx}
\usepackage{subcaption}
\usepackage{amsmath}
\usepackage{amssymb}
\usepackage{mathtools}
\usepackage[capitalize]{cleveref}
\usepackage{booktabs}
\usepackage{multirow}
\crefname{section}{Sec.}{Sec.}
\Crefname{section}{Sec.}{Sec.}
\crefname{appendix}{App.}{Apps.}
\Crefname{appendix}{App.}{Apps.}
\crefname{theorem}{Thm.}{Thms.}
\Crefname{theorem}{Thm.}{Thms.}
\crefname{proposition}{Prop.}{Props.}
\Crefname{proposition}{Prop.}{Props.}
\crefname{equation}{Eq.}{Eqs.}
\Crefname{equation}{Eq.}{Eqs.}
\crefname{table}{Tab.}{Tabs.}
\Crefname{table}{Tab.}{Tabs.}
\crefname{figure}{Fig.}{Figs.}
\Crefname{figure}{Fig.}{Figs.}
\crefname{algorithm}{Alg.}{Algs.}
\Crefname{algorithm}{Alg.}{Algs.}
\crefname{assumption}{Asm.}{Asms.}
\Crefname{assumption}{Asm.}{Asms.}
\crefname{mechanism}{Mech.}{Mechs.}
\Crefname{mechanism}{Mech.}{Mechs.}
\crefname{definition}{Def.}{Defs.}
\Crefname{definition}{Def.}{Defs.}

\newcommand{\btheta}{\bm \theta}

\newcommand{\X}{\bm X}
\newcommand{\W}{\bm W}
\newcommand{\Z}{\bm Z}
\newcommand{\G}{\bm G}

\definecolor{lightgray}{gray}{0.9}
\newcommand{\bftab}{\fontseries{b}\selectfont}

\newcommand{\nameshort}{EEP}

\title{\underline{E}fficient \underline{E}xpert \underline{P}runing for Sparse Mixture-of-Experts Language Models: Enhancing Performance and Reducing Inference Costs}

\author{%
  David S.~Hippocampus\thanks{Use footnote for providing further information
    about author (webpage, alternative address)---\emph{not} for acknowledging
    funding agencies.} \\
  Department of Computer Science\\
  Cranberry-Lemon University\\
  Pittsburgh, PA 15213 \\
  \texttt{hippo@cs.cranberry-lemon.edu} \\
}

\begin{document}

\maketitle

\begin{abstract}
The rapid advancement of large language models (LLMs) has led to architectures with billions to trillions of parameters, posing significant deployment challenges due to their substantial demands on memory, processing power, and energy consumption. Sparse Mixture-of-Experts (SMoE) architectures have emerged as a solution, activating only a subset of parameters per token, thereby achieving faster inference while maintaining performance. However, SMoE models still face limitations in broader deployment due to their large parameter counts and significant GPU memory requirements. 
In this work, we introduce a gradient-free evolutionary strategy named \underline Efficient \underline Expert \underline{P}runing (EEP) to enhance the pruning of experts in SMoE models. Specifically, EEP searches the pruning pattern and use expert merging as an memory-efficient way of fine-tuning the pruned model. EEP relies solely on model inference (i.e., no gradient computation) and achieves greater sparsity while maintaining or even improving performance on downstream tasks. EEP can be used to reduce both the total number of experts (thus saving GPU memory) and the number of active experts (thus accelerating inference).
For example, we demonstrate that pruning up to 75\% of experts in Mixtral $8\times7$B-Instruct results in a substantial reduction in parameters with minimal performance loss. 
Remarkably, we observe improved performance on certain tasks, such as a significant increase in accuracy on the SQuAD dataset (from 53.4\% to 75.4\%), when pruning half of the experts. With these results, EEP not only
lowers the barrier to deploying SMoE models,
but also challenges the conventional understanding of model pruning by showing that fewer experts can lead to better task-specific performance without any fine-tuning. Code is available at \url{https://github.com/imagination-research/EEP}.
\footnotetext{Correspondence to Yu Wang <yu-wang@mail.tsinghua.edu.cn>, Zinan Lin <zinanlin@microsoft.com>, Xuefei Ning <foxdoraame@gmail.com>.}
\end{abstract}

\section{Introduction}
Large language models have significantly advanced, evolving into highly versatile tools~\citep{llmforcomputer,few_shot,flamingo,shen2023hugginggpt,zeng2023socratic,nlp4science}. As these models grow in accordance with scaling laws~\citep{kaplan2020scaling}, the norm has shifted towards architectures with billions to trillions of parameters. However, the larger scale brings considerable deployment challenges due to increased demands on memory, processing power, and energy consumption~\citep{zhou2024survey,wan2023efficient}. In response to these challenges, there is a notable trend towards adopting sparse Mixture-of-Experts (SMoE) architectures~\citep{shazeer2017moe,switch_transformer,lepikhin2021gshard,hwang2023tutel}, as seen in models such as Mixtral $8\times7$B and $8\times 22$B~\citep{jiang2024mixtral}, Qwen1.5-MoE-A2.7B~\citep{bai2023qwen}, Qwen 2-57B-A14B~\citep{qwen2_blog}, DBRX~\citep{databricks2023}, and Grok-1~\citep{xai2024}. SMoE models activate only a subset of parameters for each token, resulting in faster inference while maintaining competitive performance compared to dense models of the same scale. For example, Mixtral $8\times7$B outperforms or matches Llama-2 70B~\citep{touvron2023llama2} and GPT-3.5 on many benchmarks, while it only activates 13B parameters to process each token. Although SMoE models have less computation per token, they remain parameter-heavy, e.g.\ Mixtral $8\times7$B has 47B parameters in total while Grok-1 reaches 314B (see \cref{tab:model_parameters} for other models). This limits their broader deployment due to the substantial GPU memory requirements. Additionally, their throughput may not be ideal as the batch size needs to be restricted to fit the model within the available device memory. Therefore, it is vital to innovate methods that can reduce the size of SMoE models without compromising their performance.

Many studies have shown that only a subset of parameters significantly contributes to performance when applying LLMs to downstream tasks~\citep{blalock2020state,kwon2022a,sajjad2023effect,xia2022structured}. Pruning is a crucial technique for eliminating redundancy in neural networks. It can be unstructured, achieving high sparsity while maintaining performance~\citep{blalock2020state,frantar23asparsegpt,sun2024a}, or structured, removing entire channels or layers to provide computational efficiency and reduced latency~\citep{ma2023llmpruner,tao2023structured,xia2022structured,dynabert,wang19eigendamage,kwon2022a}. One particularly efficient way is expert pruning in SMoE LLMs, a type of structured pruning with coarse granularity, which enhances overall efficiency. Recent expert pruning methods achieve 25\%-50\% sparsity and accelerate inference, but struggle to maintain performance~\citep{lu2024not} or need fine-tuning, requiring substantial GPU memory and resources~\citep{chen2022task,muzio2024seer}. Thus, there is a pressing need for efficient pruning methods that operate within the constraints of inference resources for SMoE LLMs. 

\begin{figure}[t!]
\vspace{-3em}
    \centering
    \begin{subfigure}[b!]{0.34\textwidth}
        \centering
        \includegraphics[width=\textwidth]{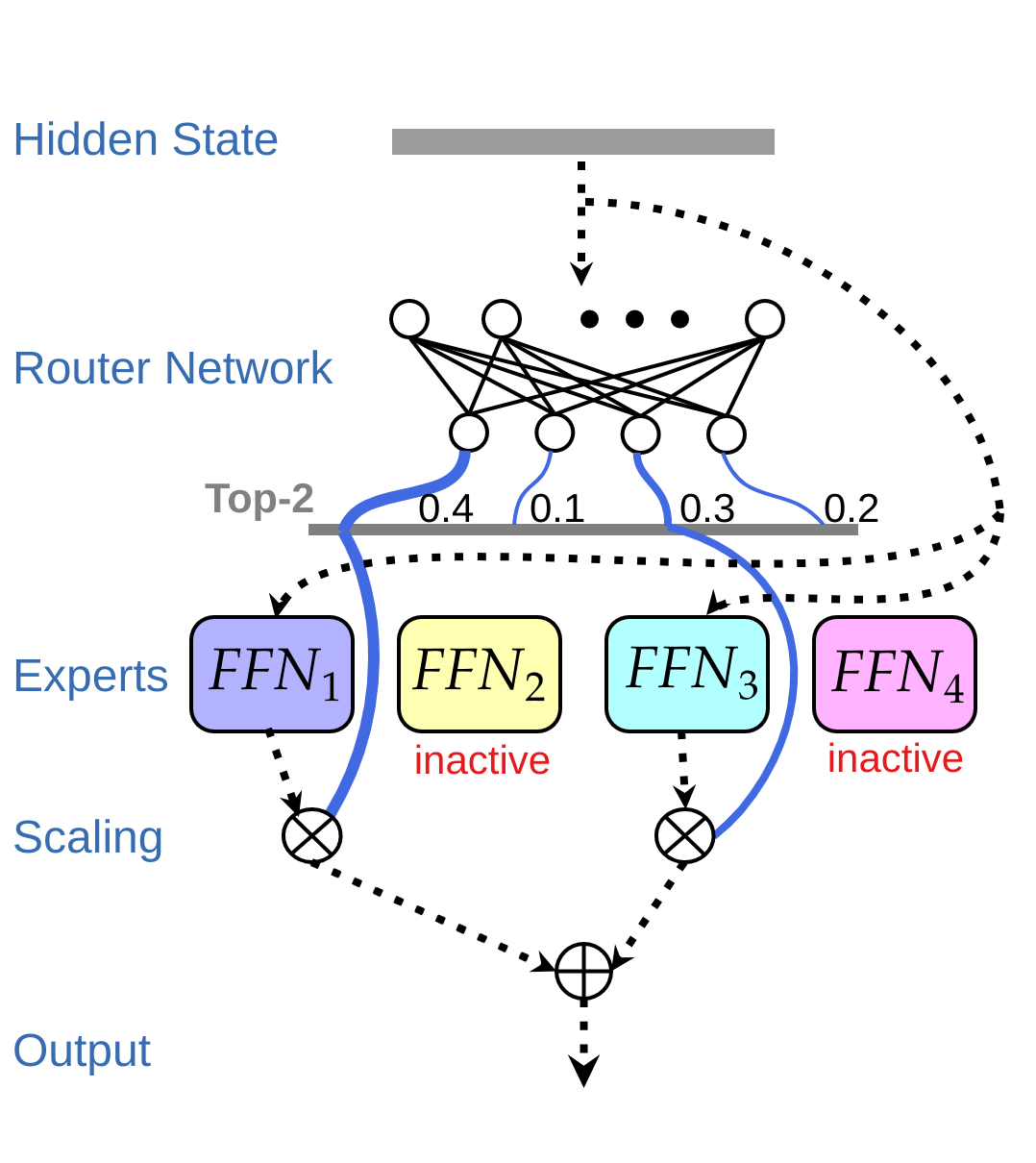} %
        \caption{A SMoE block before pruning.}
        \label{fig:illustration_baseline}
    \end{subfigure}
    \hfill
    \begin{subfigure}[b!]{0.65\textwidth}
        \centering
        \includegraphics[width=\textwidth]{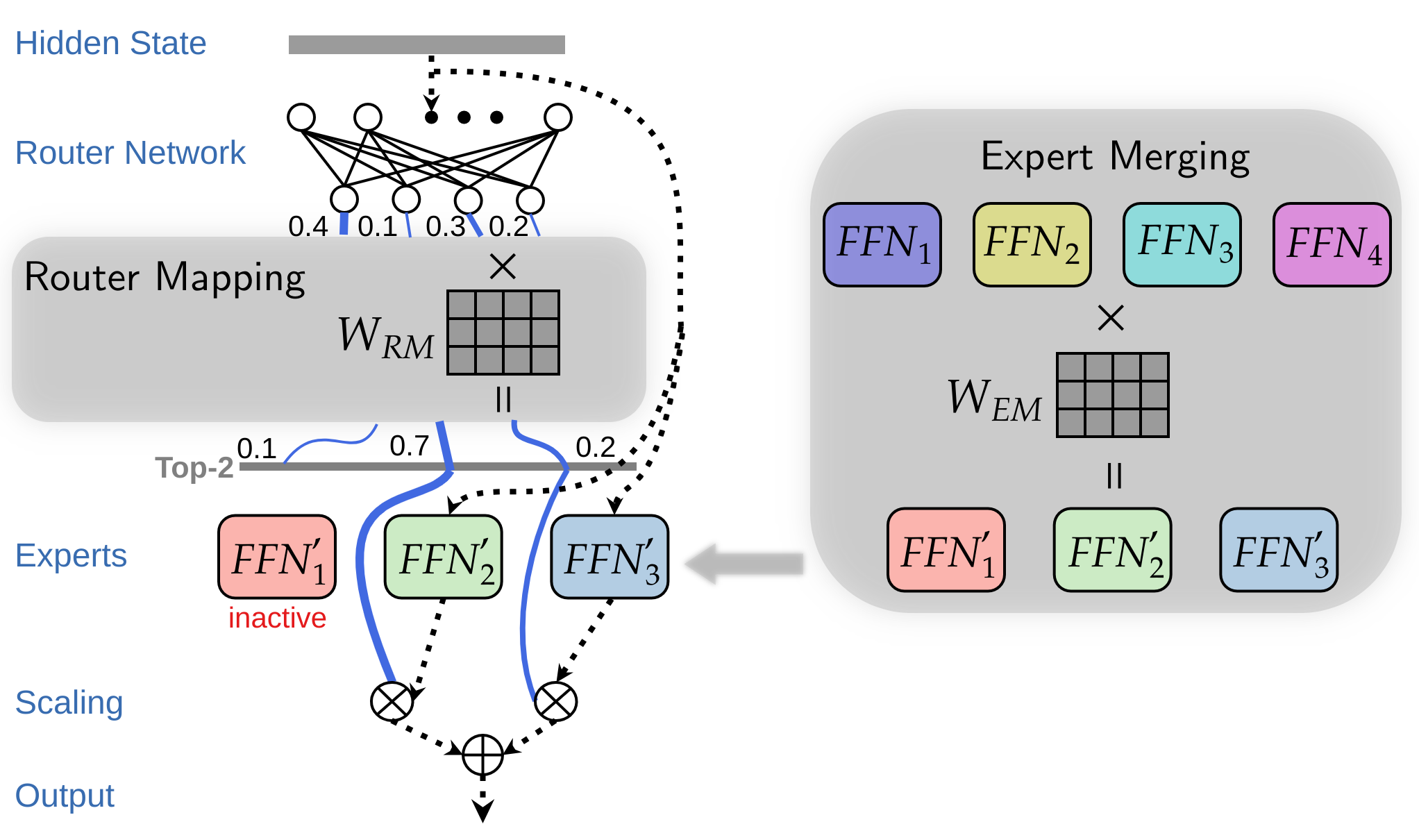} %
        \caption{Parameter space designed for expert pruning and merging.}
        \label{fig:illustration_ours}
    \end{subfigure}
    \caption{(a) the original SMoE block  and (b) our implementation of \nameshort{}. We introduce the expert merging matrix $\W_{\text{EM}}$, and the router mapping matrix $\W_{\text{RM}}$, to enable the search for the optimal pruning configuration. When $\W_{\text{EM}}$ and $\W_{\text{RM}}$ have one-hot vectors as their rows, pruning is performed. When their elements are continuous values, routing weights and experts are aggregated to generate new weights and experts. We use an evolutionary strategy to search for the optimal $\W_{\text{EM}}$ and $\W_{\text{RM}}$.}
    \label{fig:illustration}
    \vspace{-1.em}
\end{figure}

In this work, we propose a gradient-free evolutionary strategy that achieves high sparsity while maintaining performance given a small train set on the downstream tasks. Our method is divided into two phases: expert pruning and expert merging. To facilitate the search for optimal pruning configurations, we design a parameter space for router mapping and expert merging, represented by two weight matrices, $W_{RM}$ and $W_{EM}$. These matrices are applied to the router weighting and expert modules, as illustrated in \cref{fig:illustration}. In the first phase, expert pruning, we search through the weight matrices to retain the most prominent experts without updating any network parameters. In the second phase, expert merging, we retrieve knowledge from the pruned experts and consolidate it into the retained experts. To these ends, $W_{RM}$ and $W_{EM}$ are set to one-hot rows in the first phase and to real numbers in the second phase. Since our method is gradient-free, it can be conducted on devices capable of inference. Our contributions can be summarized as follows:
\begin{itemize}[leftmargin=*]
    \item \textbf{Pruning the total number of experts:  smaller memory consumption and better performance.} Our approach enables more aggressive pruning of experts compared to current methods~\citep{lu2024not,muzio2024seer}. In experiments on Mixtral $8\times 7$B-Instruct, we reduce the number of experts in each SMoE block from 8 to 2, a 72\% reduction in parameters, while maintaining comparable performance across various downstream tasks. \textit{Surprisingly, we observe that fewer experts can lead to better performance.} For instance, on the SQuAD dataset, pruning 4 out of 8 experts result in a performance increase from 53.4\% to 75.4\% without updating the remaining experts.
    \item \textbf{Pruning the number of \textit{active} experts: better inference efficiency.} We explore the pruning of active experts and find that effective expert merging compensates for the loss of active experts across downstream tasks. This process significantly improves efficiency without compromising the model's utility on these tasks. For instance, by reducing the active experts in Mixtral 8 $\times$ 7B from two to one, we observe a prefill acceleration of up to 1.63$\times$.
    \item \textbf{Generalization ability.} We test the performance of our method on datasets with higher diversity and out-of-distribution tasks using MMLU~\citep{hendryckstest2021}. Specifically, we take 50 of the 57 datasets included in MMLU and conduct \nameshort{} using data from a small subset of each of the 50 datasets. We then evaluate the pruned model on the test data of \textbf{i}) the 50 datasets and \textbf{ii}) the 7 unseen datasets. In both evaluation tasks, we observe that EEP consistently outperforms other pruning methods, demonstrating the strong generalization ability of our method.
    \item \textbf{A novel and efficient pruning paradigm.} 
    Common pruning paradigm usually conducts two steps. In the first step, parameters are pruned using empirical criteria. This operation often lowers performance. In the second step, retained parameters are fine-tuned through stochastic gradient descent to recover performance. This operation often requires substantial GPU memory and computation time, making it prohibitive for most users of LLMs. \nameshort{} adopts a gradient-free evolutionary strategy for both pruning and fine-tuning. As a result, our pruned model significantly outperforms the pruned models of other methods, while our pruning and fine-tuning processes can run on devices affordable for inference, making \nameshort{} more widely applicable.
    In addition, to inherit knowledge from the unpruned model, existing methods either select a subset of weights based on predefined importance criteria~\cite{he2018amc,yang2018netadapt,ning2020dsa}, or rely on distillation techniques~\cite{polino2018model,selfdistill,aghli2021combining}. In contrast, \nameshort{} introduces a novel approach as a third paradigm, employing weight merging~\citep{wortsman22aModelsoup} to transfer knowledge during the model compression process.
  
\end{itemize}

\vspace{-1em}
\section{Related work}
\vspace{-0.5em}
\textbf{Sparse Mixture-of-Experts LLMs.}
Shazeer et al.\ \citep{shazeer2017moe} introduced the sparse MoE layer, which consists of multiple experts, each being a simple feed-forward network (FFN), and a trainable router network that selects a sparse combination of the experts to process each input. Such SMoE models can significantly increase model capacity while maintaining computational efficiency. However, this utility is ideally achieved when the router accurately and evenly assigns experts to each token during training and inference. Many works address these challenges~\citep{switch_transformer,pmlrlewis21abase,daistablemoe,NEURIPS2022expertchoice}.  Recently, many SOTA LLMs adopt the SMoE structure to achieve high performance and computational efficiency simultaneously~\citep{jiang2024mixtral,bai2023qwen,databricks2023,xai2024}. Additionally, Zhang et al.\ \citep{zhangmoefication} propose transforming non-MoE models into SMoE models to accelerate inference, and Komatsuzaki et al.\ \citep{komatsuzaki2023sparse} upcycle pretrained models by reusing the parameters to initialize SMoE models, where all experts are replicates of the original FFNs, and then fine-tune the SMoE models.

\textbf{Pruning for LLMs.}
Pruning techniques have emerged as a crucial strategy for optimizing LLMs by reducing model size and computational costs while maintaining performance. Unstructured pruning~\citep{blalock2020state,frantar23asparsegpt,sun2024a,syed2023prune} entails the removal of individual weights according to specific criteria, creating sparse networks that demand specialized hardware for efficient execution. In contrast, structured pruning~\citep{ma2023llmpruner,tao2023structured,xia2022structured,dynabert,wang19eigendamage,kwon2022a,child2019generating,xiao2024efficient,beltagy2020longformer} eliminates entire structures, such as neurons or attention heads, facilitating more straightforward implementation on standard hardware. Within structured pruning, specific focus areas include attention mechanisms, where redundant heads are pruned to streamline the self-attention layers, and FFNs where unnecessary neurons are removed to enhance computational efficiency. Additionally, expert pruning for SMoE models selectively prunes the expert networks~\citep{lu2024not,muzio2024seer,chen2022task,koishekenov2023memory}.

\textbf{Evolutionary Strategy for Optimization.}
Evolutionary Strategies (ES) have been increasingly recognized for their robustness and flexibility in various optimization tasks, particularly where gradient-based methods fall short~\cite{wierstra14aNaturalEvolution}. Notably, ES is highly effective for optimizing non-differentiable objective functions, offering a powerful alternative in scenarios where gradients are unavailable or unreliable~\citep{salimans2017evolution,FedLES,liu2024linear,trofin2021mlgo,liu2023oms}. Furthermore, ES excels in discrete optimization spaces, making it suitable for a wide range of combinatorial problems~\citep{akiba2024evolutionary,liu2024a,liu23OpsDpm}. Recent advancements have extended the application of ES to the domain of LLMs, enabling memory-efficient fine-tuning without the need for backpropagation~\citep{malladi2023finetuning}. 

\begin{figure}[t!]
  \centering
  \begin{minipage}{.75\textwidth} %
    \centering
    \includegraphics[width=\linewidth]{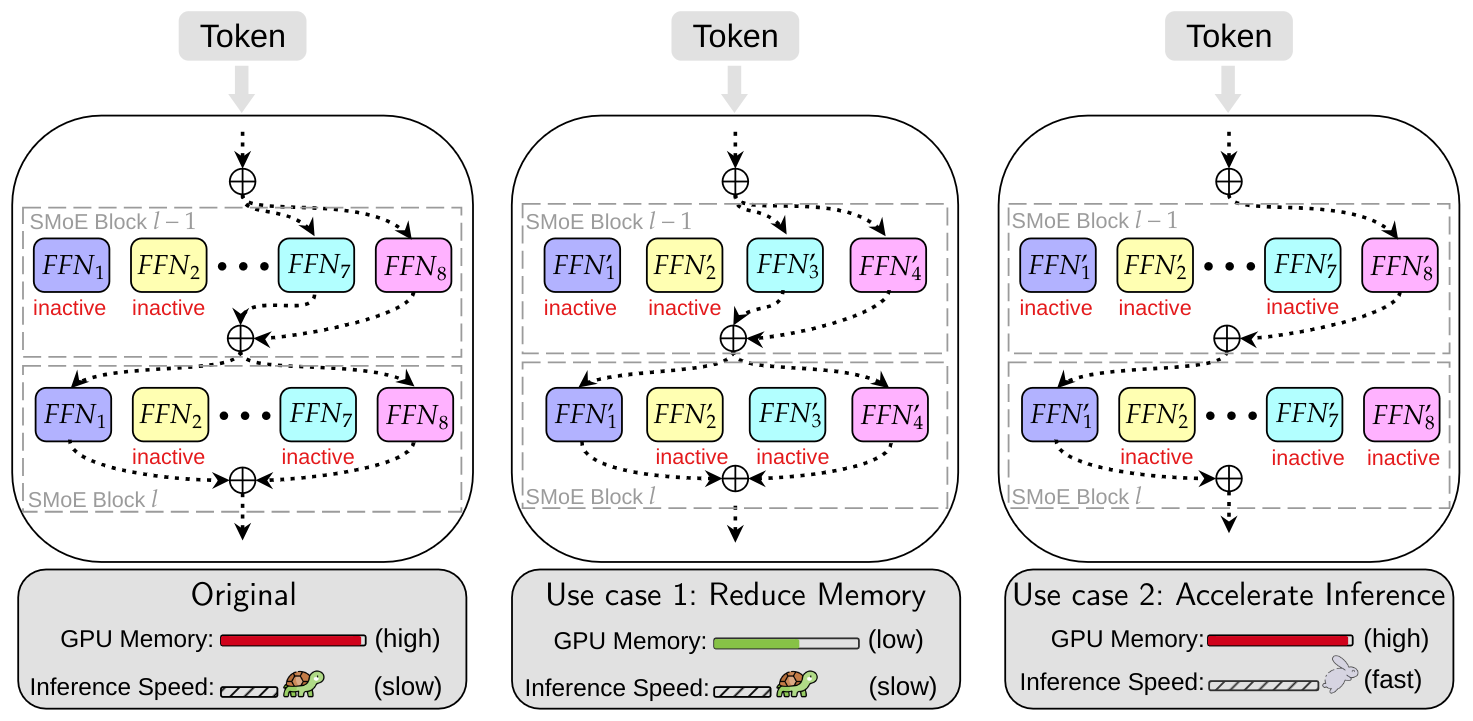}
  \end{minipage}%
  \hfill
  \begin{minipage}{.24\textwidth} %
    \caption{We leverage \nameshort{} for two purposes: reducing the total number of experts, which lowers the memory footprint (use case 1), and reducing the number of active experts, thereby accelerating inference (use case 2).}
  \label{fig:use_case}
  \end{minipage}
    \vspace{-1em}
\end{figure}

\section{Background of sparse Mixture-of-Expert language model}
In this section, we discuss the general concept of sparse Mixture-of-Experts (SMoE) implementation in modern decoder-only models, using the Mixtral family~\citep{jiang2024mixtral} as a specific focus. A schematic illustration is provided in \cref{fig:illustration_baseline}.

\textbf{Notations.}
Let $\X \in \mathbb{R}^{n \times d}$ represent the input to a SMoE block, where $n$ is the sequence length and $d$ is the hidden dimension. The output of the attention block is denoted by $\Z \in \mathbb{R}^{n \times d}$. The main parameters in the attention block are the weight matrices for computing query, key, and value: $\W_Q, \W_K, \W_V$. In the SMoE structure, there are $E$ experts, each represented by a feed-forward network (FFN) with parameters $\btheta_i$ for the $i$-th expert. The router network, denoted by $\W_R$, produces routing weights $\G \in \mathbb{R}^{n \times E}$ for the sparse activation of the experts. For clarity, we omit the normalization layers and biases.

\textbf{Self-Attention Mechanism.}
The self-attention mechanism computes the query, key, and value matrices as follows: $\bm Q = \X \W_Q,\, \bm K = \X \W_K,\, \bm V = \X \W_V$. The attention scores and the output $\Z$ are then computed as:
\begin{align}
\text{Attention}(\bm Q, \bm K, \bm V) = \text{softmax}\left(\frac{\bm Q \bm K^\top}{\sqrt{d_k}}\right) \bm V,\quad \Z = \text{Attention}(\bm Q, \bm K, \bm V)\W_O,
\end{align}
where $\text{softmax}(\cdot)$ denotes a row-wise softmax function.
The attention mechanism produces a weighted sum of the values $\bm V$, where the weights are derived from the dot product of the queries $\bm Q$ and keys $\bm K$, scaled by the square root of  key/query dimension $\sqrt{d_k}$. Then the weighted averaged values are mapped by the output matrix $\W_O$ to $\Z$.

\paragraph{Router Network in SMoE Structure.}
The router network determines which experts to activate and how to scale their outputs. The routing weights $\G \in \mathbb{R}^{n \times E}$ are computed as:
\begin{align}
\G &= \text{softmax}(\Z \W_G).
\end{align}
Sparse activation of the experts is achieved by selecting the top-$k$ routing weights for each input token. The output of the activated experts is scaled by the routing weights and aggregated to form the output of the SMoE layer $\bm{H}$:\footnote{The top-$k$ routing weights may be further normalized to sum to 1; this nuance is omitted here.}
\begin{align}
\forall j=1\dots n, \quad \bm{H}_j = \sum_{i \in \text{TopK}(\G_j)} \G_{ji} \cdot \text{FFN}_i(\Z_j),
\end{align}
where $\text{TopK}(\G_j)$ denotes the indices of the top-$k$ routing weights for the $j$-th input token, and $\text{FFN}_i$ denotes the function of the $i$-th expert, as defined below.

\textbf{FFN as Expert.}
Each expert in the SMoE structure is an independent FFN with two fully-connected layers, denoted by $\W_{1i}$ and $\W_{2i}$. When applying SwiGLU~\citep{shazeer2020glu}, an additional weight matrix $\W_{3i}$ is introduced for the activation function. The $i$-th expert processes the input as follows:
\begin{align}
\label{eq:ffn}
\text{FFN}_i(\Z_{sub}) &= \text{SwiGLU}(\Z_{sub}, \W_{1i}, \W_{3i}) \W_{2i},
\end{align}
where $\Z_{sub}$ denotes the a subset of rows in $\Z$ that activates the $i$-th expert.
Depending on the activation function, the parameters of the $i$-th expert are either $\btheta_i = \{\W_{1i}, \W_{2i}\}$ or $\btheta_i = \{\W_{1i}, \W_{2i}, \W_{3i}\}$.

\section{Method}
\begin{wrapfigure}{R}{0.4\textwidth}
  \centering
  \includegraphics[width=\linewidth]{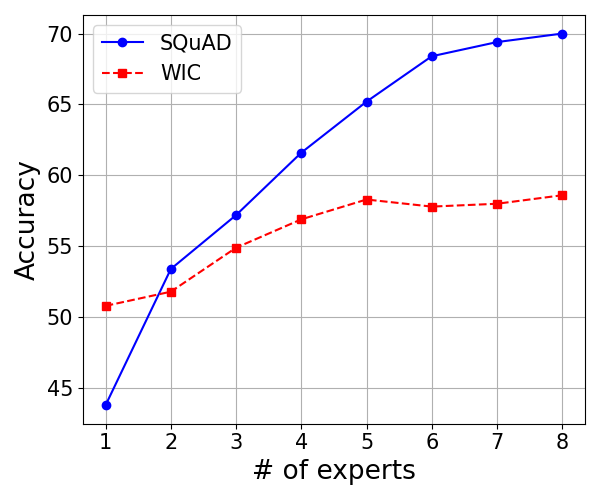} %
\vspace{-2em}
  \caption{Performance from a single expert to an ensemble of experts.}
  \label{fig:motivation}
  \vspace{-1.5em}
\end{wrapfigure}
In this section, we introduce our proposed approach for optimizing SMoE LLMs through expert pruning and merging. We aim to enhance the efficiency and performance of SMoE architectures by leveraging evolutionary strategies. Our method addresses the challenges of large and complex search spaces without incurring the prohibitive computational costs associated with gradient-based optimization. The subsequent subsections elaborate on our motivation (\cref{sec:method:motivation}), the configuration of the  parameter space (\cref{sec:method:parameter}), the evolutionary optimization strategy employed to achieve our objectives (\cref{sec:method:es}), and the use cases we apply \nameshort{} (\cref{sec:usecase}).

\subsection{Motivation}
\label{sec:method:motivation}
LLMs based on the SMoE architecture have shown remarkable performance across various natural language processing tasks~\citep{jiang2024mixtral,databricks2023,xai2024}. These models leverage multiple experts, activating only a subset for any given input, thus balancing computational efficiency and model capacity. Typically, top-2 experts are activated, striking a balance between performance and computational cost.

\Cref{fig:motivation} presents our investigation into the activation of different numbers of experts on Mixtral $8\times 7$B-Instruct, revealing the following observations: i) Activating only a single expert does not lead to model collapse and may result in only a minimal performance drop compared to the default setting of using two experts. This suggests that individual experts possess redundant knowledge, enabling them to maintain reasonable performance independently. This redundancy indicates potential for expert pruning. ii) Conversely, activating all 8 experts leads to a noticeable performance gain, highlighting the benefits of expert ensemble. However, the computational cost of such an ensemble is substantially higher. Wortsman et al.\ \citep{wortsman22aModelsoup} have shown that merging differently fine-tuned models can efficiently substitute their ensemble, achieving similar performance with reduced computational overhead.

Building on these insights, we propose a two-step approach involving expert pruning followed by expert merging. Initially, we search for the optimal subset of experts given a fixed size. Subsequently, we employ expert merging to consolidate the knowledge from the pruned experts into the remaining ones. This approach not only restores the knowledge of the pruned experts but also updates the surviving experts to incorporate the collective expertise of the entire SMoE block.

\subsection{Parameter space for expert pruning and merging}
\label{sec:method:parameter}
\textbf{Expert Pruning and Merging Matrices.}
To efficiently prune and merge experts in each SMoE block ($l=1 \dots L$), we introduce two key matrices: the Router Mapping matrix ($\bm{W}_{\text{RM}}^l$) and the Expert Merging matrix ($\bm{W}_{\text{EM}}^l$). For clarity, we omit the block index $l$ in this section. A schematic illustration is provided in \cref{fig:illustration_ours}. The router mapping matrix $\bm{W}_{\text{RM}} \in \mathbb{R}^{E' \times E}$, where $E'$ is the reduced number of experts (i.e.,\ $E$ > $E'$), is applied to the routing weights $\bm{G}$ to reduce the dimensionality and handle fewer experts:
\begin{align}
\bm{G}' = \bm{W}_{\text{RM}}\text{softmax}(\Z \W_G),
\end{align}

The expert merging matrix $\bm{W}_{\text{EM}} \in \mathbb{R}^{E' \times E}$ is applied to the expert weights $\{\btheta_i\}_{i=1}^E$ to merge $E$ experts into $E'$ experts. Each element in $\bm{W}_{\text{EM}}$ operates blockwise on the parameters of the experts. Denote $\{\omega_{j1}, \omega_{j2}, \dots, \omega_{jE}\}$ as the $j$-th row of $\bm{W}_{\text{EM}}$ that maps the original $E$ experts to the $j$-th new expert $\btheta'_j$. We define merging as follows:
\begin{align}
\btheta'_j = \{\sum_{i=1}^E \omega_{ji} \W_{1i}, \sum_{i=1}^E \omega_{ji} \W_{2i}, \sum_{i=1}^E \omega_{ji} \W_{3i}\},
\end{align}
where the parameters of the experts are defined in \Cref{eq:ffn}.

\textbf{Expert Pruning Phase.}
During the expert pruning phase, the low-rank matrices $\bm{W}_{\text{RM}}$ and $\bm{W}_{\text{EM}}$ are initialized with each row as a one-hot vector to ensure that only pruning occurs. Additionally, $\bm{W}_{\text{RM}}$ and $\bm{W}_{\text{EM}}$ are set as to be identical $\bm{W}_{\text{RM}} = \bm{W}_{\text{EM}}$. Consequently, these matrices only retain the selected expert weights and their corresponding routing weights. During evolutionary search, \nameshort{} also maintains the one-hot format of $\bm{W}_{\text{RM}}$ and $\bm{W}_{\text{EM}}$.

\textbf{Expert Merging Phase.}
In the expert merging phase, $\bm{W}_{\text{RM}}$ and $\bm{W}_{\text{EM}}$ are decoupled and initialized from their optimal values obtained during the pruning phase. This decoupling allows for a more flexible transformation where multiple experts can be merged, and the router weights can be updated independently. During this phase, the elements of $\bm{W}_{\text{RM}}$ and $\bm{W}_{\text{EM}}$ transition from discrete $0/1$ values to continuous values. This allows the matrices to perform more nuanced transformations.

\subsection{Evolutionary search for the router mapping and expert merging matrices}
\label{sec:method:es}
The search space of the router mapping and expert merging matrices is large and complex, making it difficult to design heuristics for determining a solution, as is done in other expert pruning studies~\citep{muzio2024seer,chen2022task,lu2024not}. Therefore, an efficient optimization strategy is necessary. Given the substantial size of SMoE LLMs, computing gradients for optimization is computationally prohibitive for most users. As a solution, we employ a gradient-free evolutionary strategy, similar to approaches found in previous works~\cite{liu23OpsDpm,liu2024linear}. Our algorithm is detailed in \cref{alg:evo_search}.

Initially, we populate the search space using random initialization. During the evolutionary search, each set of router mapping and expert merging matrices is treated as an individual. In each iteration, only the top-performing individuals are selected as parents to produce the next generation through crossover and mutation. Specifically, during crossover, we randomly combine the entries of the matrices from two parents or select one parent's matrices entirely. For mutation, we introduce random Gaussian noise to the matrices, ensuring stochastic variations. This process conserves beneficial adaptations while discarding detrimental modifications, enhancing the optimization process.

This evolutionary reproduction process is repeated for a predetermined number of iterations within each search phase, updating the population with newly generated individuals. Upon completion of the search process, the best individual is selected as the output of our search algorithm. 

\subsection{Use Cases}
\label{sec:usecase}
We explore two applications of \nameshort{}: expert pruning and expert activation pruning. In expert pruning, \nameshort{} searches for optimal router mapping ($\W_\text{RM}$) and expert merging matrices ($\W_\text{EM}$) to minimize the total number of experts while maintaining high performance. For expert activation pruning, the goal is to achieve strong performance with only one active expert per token. Here, we use the same \nameshort{} search algorithm to conduct expert and router networks optimization by updating the $\W_\text{RM}$ and $\W_\text{EM}$ matrices, while only activates one expert during inference. \Cref{fig:use_case} illustrates these two use cases. Additionally, we investigate the combination of these two approaches, reducing both the total number of experts and the number of active experts simultaneously (see \cref{sec:active_number}).

\section{Experiments}
\label{sec:exp}
In this section, we validate the effectiveness of our method by considering two use cases: expert pruning and expert activation pruning. In \cref{sec:exp_setting}, we introduce the experimental settings. In \cref{sec:total_number}, we investigate the first use case, expert pruning, by applying \nameshort{} to reduce the total number of experts. In \cref{sec:active_number}, we further explore expert activation pruning, applying \nameshort{} to maintain performance while reducing the number of active experts by changing the top-2 routing weights to top-1. We also examine a composite case where both the total number of experts and the number of active experts are reduced. In \cref{sec:larger_dataset}, we present the experimental results on larger and more diverse datasets, as well as performance on out-of-distribution datasets, to validate the generalization ability of \nameshort{}. In \cref{sec:profile}, we profile memory usage and inference speed to demonstrate that our method achieves significant improvements compared to the full SMoE models. In \cref{sec:analysis} we provide insights on the observation of fewer experts but higher performance. More results, including experiments on larger datasets and other models, can be found in \cref{app:res}.

\subsection{Experimental settings}
\label{sec:exp_setting}
Our main results are based on the popular SMoE models Mixtral 8$\times$7B~\cite{jiang2024mixtral}. We also include a larger model, Mixtral 8$\times$22B~\cite{jiang2024mixtral}, to demonstrate the generalization of our methods. We use the "Instruct" version of these models for the generation tasks. We select tasks from the SuperGLUE dataset, as well as several other generation tasks, including SQuAD~\cite{rajpurkar2016squad} and DROP~\citep{dua2019drop}.
For each individual dataset, we randomly sample a subset from the training set to conduct evolutionary search and use the test set for evaluation. Additional details can be found in \cref{app:exp}.

\textbf{Evaluation.} We adopt a generation-based evaluation approach for all datasets. Specifically, we use the instruction fine-tuned model to generate answers directly in response to the given questions and apply template matching to determine the correctness of the answers. Our evaluation protocol primarily follows the implementation of OpenCompass \citep{2023opencompass} for the design of question prompts, types of templates, and matching criteria, with a few modifications to better suit the Mixtral family of models. All experiments use the same evaluation settings. Examples of prompts and model outputs can be found in \cref{app:prompt} and \cref{app:example}.

\textbf{Baselines.} Since our method aims to compress the instruction fine-tuned SMoE models on downstream tasks, we consider the zero-shot performance as our main baseline to show that \nameshort{} can achieve a significant decrease on the memory footprint and/or computation overhead during the inference time while maintain or even achieve better performance. For the use case of decreasing the total number of experts, we additionally compare \nameshort{} with four other types of baseline to demonstrate the effectiveness of the designed search space and the evolutionary-search-based tuning method: \textbf{(1)} \verb|Random| selection of pruned experts, \textbf{(2\&3)} Pruning the experts with the lowest \verb|frequency| of being activated or the lowest \verb|soft| activation values~\citep{muzio2024seer}, and \textbf{(4)} \verb|NAEE| \cite{lu2024not}, which exhaustively evaluates the loss between the full model and all pruning choices for each layer and select the one with the lowest loss. For the use case of decreasing the active number of experts, we select the dynamic skipping method proposed by \verb|NAEE| \cite{lu2024not} as an additional baseline. More details are given in \cref{app:exp}.

\subsection{Reducing the total number of experts}
\label{sec:total_number}
We apply \nameshort{} to search for the optimal pruning configuration, parameterized by the router mapping matrix $\W_{\text{RM}}$ and the expert merging matrix $\W_{\text{EM}}$, for maintaining 4 experts and 2 experts. \nameshort{} (Prune Only) indicates the results from solely conducting the expert pruning phase as described in \cref{sec:method:parameter}. In contrast, \nameshort{} (Prune + Merge) shows the results after the complete evolutionary search process. The results are shown in \cref{tab:8x7_total_num}, and we discuss them below. \verb|Random| is conducted 30 times, and we present the mean results here, deferring the complete results to \cref{app:random_search}.
\begin{table*}[t]
\centering
    \caption{Results of expert pruning on Mixtral 8$\times$7B-Instruct. \textbf{Bold} values indicate the best across all methods; \underline{underlined} values show the best without parameter updates (i.e., excluding \nameshort{} (Prune+Merge)).}
      \resizebox{\columnwidth}{!}{%
    \begin{tabular}{cccccccccccc|c}
      \toprule
         Expert & Method & COPA & MultiRC & WIC & WSC & RTE & BoolQ & CB & ReCoRD & DROP & SQuAD &Avg. \\
      \midrule
         Num=8 & Full Model & 89.0 & 83.0 & 51.8 & 63.5 & 73.2 & 77.4 & 51.7 & 50.3 & 30.6 & 53.4& 62.4\\
      \midrule
         \multirow{6}{*}{Num=4}& Random & 63.8 & 49.4 & 37.6 & 43.3 & 45.1 & 50.2 & 38.7& 35.1& 27.4& 58.3& 44.9\\
         & Frequency~\citep{muzio2024seer} & 63.0 & 74.8 & 36.0& 34.6& 18.1& 71.0& 30.4& 41.6& 29.9& 58.2& 45.8\\
         & Soft Activation~\citep{muzio2024seer} & 73.0 & 30.6 & 51.4 & 37.5 & 41.9 & 40.4 & 17.9 & 36.8 & 33.3 & 10.2 & 37.3 \\
         & NAEE~\citep{lu2024not} & 87.0 & 76.0 & 52.6& 64.5& 61.7& 77.2& 51.7& 50.4& 30.6& 53.0& 60.5\\
      \cmidrule{2-13}
         & \nameshort{} (Prune Only) & \underline{95.0} & \underline{81.2} & \underline{57.8} & \underline{67.3} & \underline{74.0} & \underline{82.8} & \underline{69.6} & \underline{60.0} & \underline{37.3} & \underline{75.2}& \underline{70.3}\\
         & \nameshort{} (Prune+Merge) & \bftab 99.0 & \bftab 84.6 & \bftab 65.0 & \bftab 73.1 &\bftab  76.9 & \bftab 84.8 &\bftab 75.0 &\bftab 63.6 &\bftab 39.7 &\bftab 80.6 & \bftab 74.2\\
       \midrule
         \multirow{6}{*}{Num=2}& Random & 36.8 & 22.3 & 13.6& 15.0& 28.4& 15.5& 38.6&16.9& 18.3& 36.9& 24.2\\
         & Frequency~\citep{muzio2024seer} & 51.0 & 17.6 & 8.8& 1.9& 48.4& 30.6& 35.7& 10.4& 14.9& 9.2& 24.9\\
         & Soft Activation~\citep{muzio2024seer} & 33.0 & 18.2 &49.4 & 18.5 & 15.2 & 1.8 & 32.1 & 4.4 & 11.7 & 50.0 & 23.4 \\
         & NAEE~\citep{lu2024not} & 75.0 & 42.4 & 48.4& 49.0& 54.5& 49.8& 19.6& 42.0& 31.2& 58.2& 47.0\\
       \cmidrule{2-13}
         & \nameshort{} (Prune Only) & \underline{76.0} & \underline{63.8} & \underline{51.8}& \underline{63.5}& \underline{64.3}& \underline{70.6}& \underline{58.9}& \underline{47.2}& \underline{37.1}& \underline{64.0}& \underline{59.7}\\
         & \nameshort{} (Prune+Merge) & \bftab 93.0 & \bftab 71.6 & \bftab 58.6&\bftab 65.4&\bftab 69.0&\bftab 75.6&\bftab 66.1&\bftab 47.2&\bftab 38.4&\bftab  70.2& \bftab 65.6\\
      \bottomrule
    \end{tabular}
    }
  \label{tab:8x7_total_num}
\end{table*}

\textbf{\nameshort{} fully exploits expert-wise redundancy on downstream tasks}. Based on the results obtained from the pruning phase of \nameshort{}, retaining only 4 experts allows the model to achieve better performance and lower computational costs simultaneously on most datasets, except for MultiRC. Even with a particularly low budget of retaining only 2 experts, \nameshort{} can still achieve comparable or even better performance than the full model on five datasets, with some datasets showing significant improvements over the best baseline (e.g., 58.9 vs. 51.7 on CB and 64.0 vs. 53.4 on SQuAD). For the remaining datasets, model collapse is avoided.

\textbf{\nameshort{} is more effective than other baseline methods for selecting pruned experts.} Comparing the results of other methods, we find that \nameshort{} is more effective for identifying the optimal pruning pattern. Random sampling of experts results in low mean accuracy and high variance. Pruning experts based on selection frequency also performs poorly on most datasets and has a high probability of collapse under high sparsity. \verb|NAEE| can nearly maintain the performance of the full model when retaining four experts. However, \nameshort{} surpasses all methods by a large margin across all datasets.

\textbf{Expert merging brings significant improvements after pruning}. As shown in the last row for each pruning rate in \cref{tab:8x7_total_num}, the results after expert merging exceed those obtained through the expert pruning phase alone. Specifically, expert merging achieves a general improvement on almost all datasets. On WIC, CB, and SQuAD under both pruning rates, and on WSC when four experts are retained, the accuracy improvement reaches 5\%$\sim$7\%, demonstrating its effectiveness in restoring the knowledge of pruned experts and enhancing individual experts. Additionally, we find expert merging to be an effective method for fine-tuning SMoE LLMs (i.e., keeping the number of total and active experts); the results of this are presented in \cref{tab:fine-tune}.

\textbf{Generality across models.} With the promising results of Mixtral 8$\times$7B-Instruct model, we further apply \nameshort{} to a larger model: Mixtral 8$\times$22B-Instruct \cite{jiang2024mixtral}, Qwen1.5-MoE-A2.7B-Chat \cite{bai2023qwen}, and Qwen2-MoE-A14B-Chat \cite{qwen2_blog}. We conduct experiments on fewer datasets due to the constraint of computational resource. Results are shown at \cref{tab:8x22_total_num}, \cref{tab:qwen1.5_total_num}, and \cref{tab:qwen2_total_num}, respectively. \nameshort{} also achieves a strong improvement and above observations are still held, which indicates the scaling-up ability of \nameshort{} towards large SMoE models.

\begin{table*}[t]
\centering
    \caption{Results of expert pruning on Mixtral $8\times 22$B-Instruct. \textbf{Bold} values indicate the best across all methods; \underline{underlined} values show the best without parameter updates (i.e., excluding \nameshort{} (Prune+Merge)).}
    \begin{small}
    \begin{tabular}{ccccccc|c}
      \toprule
         Budget & Method & WIC & WSC & BoolQ & CB & SQuAD & Avg.\\
      \midrule
         Num=8 & Full Model & 68.2 & 81.7 & 90.2 & 46.5 & 45.8 & 66.5 \\
      \midrule
         \multirow{7}{*}{Num=4}& Random & 27.0 & 30.2 & 37.8 & 34.6 & 37.2 & 33.4 \\
         &Frequency~\citep{muzio2024seer} & 0.0 & 38.5 & 76.6 & 57.1 & 50.6 & 30.6\\
         &Soft Activation~\citep{muzio2024seer} & 25.2 & 60.6 & 6.4 & 60.7 & 54.2 & 41.4 \\
         & NAEE~\citep{lu2024not} & 64.0 & 68.3 & 78.4 & 33.9 & 52.4 & 59.4 \\
      \cmidrule{2-8}
         & \nameshort{} (Prune Only) & \underline{70.2} & \underline{84.2} &\underline{\bftab 89.6} & \underline{75.0} & \underline{71.4} & \underline{78.1}\\
         & \nameshort{} (Prune+Merge) &\bftab 72.2 &\bftab 87.5 &\bftab 89.6 &\bftab 78.6 &\bftab 74.0 & \bftab 80.4 \\
       \midrule
         \multirow{7}{*}{Num=2}& Random & 13.9 & 10.1 & 11.0 & 24.9 & 15.6 & 15.1\\
         &Frequency~\citep{muzio2024seer} & 0.0 & 0.0 & 0.0 & 0.0 & 0.0 & 0.0\\
         &Soft Activation~\citep{muzio2024seer} & 2.4 & 1.9 & 3.6 & 19.6 & 52.6 & 16.0\\
         & NAEE~\citep{lu2024not} & 34.0 & 32.7 & 45.0 & 16.1 & 50.0 & 30.6 \\
       \cmidrule{2-8}
         & \nameshort{} (Prune Only) & \underline{57.8} & \underline{63.5} & \underline{76.0} & \underline{50.0} & \underline{71.0} & \underline{63.7}\\
         & \nameshort{} (Prune+Merge) & \bftab 59.6 &\bftab 65.4 &\bftab 76.4 &\bftab 58.9 &\bftab 75.0 & \bftab 67.1\\
      \bottomrule
    \end{tabular}
    \end{small}
  \label{tab:8x22_total_num}
\end{table*}

\subsection{Reducing the number of active experts}
\label{sec:active_number}
Next, we present the experimental results for the second use case: decreasing the number of active experts. We modify the number of active experts by changing the top-k from $k=2$ to $1$ while applying \nameshort{} to restore model performance. We evaluate our method with two different total numbers of experts (8 and 4). The results are presented in \cref{tab:8x7_active_num}. We summarize the observations below.

\textbf{\nameshort{} can improve individual experts through expert merging, allowing a single expert to handle the inference.} Keeping the total number of experts at 8 and reducing the number of active experts to 1 consistently leads to a decline in baseline performance. However, by optimizing the model with \nameshort{}, we introduce a reliable improvement that mitigates this gap, resulting in comparable or even better performance than the full model. It is important to note that when the total number of experts is maintained, there is no expert pruning phase; only expert merging is applied for \nameshort{}.

\textbf{The two use cases can be combined through \nameshort{}.} By retaining fewer experts and simultaneously reducing the number of active experts, we achieve significant savings \emph{in both GPU memory and inference time} (see \cref{sec:profile}). \nameshort{} can be directly applied in this scenario. Results show that with 4 total experts and 1 active expert, \nameshort{} achieves performance comparable to or even better than the full model.

\begin{table*}[t]
\centering
    \caption{Results of active expert pruning on Mixtral $8\times 7$B. Bold values show the best performance. ``Active'' indicates the average number of experts active per token. Avg.\ stands for average.}
    \begin{small}
    \begin{tabular}{cccccccc|c}
      \toprule
         Total & Active & Method & WIC & WSC & BoolQ & CB & SQuAD & Avg.\\
      \midrule
         \multirow{3}{*}{8} & 2 & Full Model & 51.8& 63.5& 77.4& 51.7& 53.4 &59.6\\
      \cmidrule{2-9}
         & 1 & Full Model & 50.8& 48.1& 66.0& 48.2 & 43.8& 51.4 \\
         & 1.4$\sim$1.5 & Dyn~\citep{lu2024not} & 50.0& 59.6& 72.8& 46.4  &44.8&54.7 \\
      \cmidrule{2-9}
         & 1 & \nameshort{} & \bftab 59.2&\bftab 70.2 & \bftab 79.0&\bftab 66.1&\bftab 51.8&\bftab 65.3\\
       \midrule
         \multirow{4}{*}{4} & 1 & NAEE~\citep{lu2024not} &48.6 & 20.2& 56.2& 33.9 & 51.8& 42.1\\
         & 1.4$\sim$1.5 & NAEE+Dyn~\citep{lu2024not} & 43.4& 61.5& 36.2& 53.6 & 53.4& 49.6\\
       \cmidrule{2-9}
         & 1 & \nameshort{} &\bftab 55.8 &\bftab 70.2 &\bftab 74.4&\bftab 64.3  &\bftab 72.0 &\bftab 67.3\\
      \bottomrule
    \end{tabular}
    \end{small}
  \label{tab:8x7_active_num}
\end{table*}

\subsection{In-distribution and out-of-distribution generalization on diverse datasets}
\label{sec:larger_dataset}
In this section, we further test \nameshort{} on a larger dataset, MMLU, to validate the generalization ability of \nameshort{}. We randomly split all 57 datasets in MMLU into two subsets containing 50 and 7 datasets, as the base dataset and the out-of-distribution (OOD) test dataset, respectively. We further divide each dataset in the larger subset into training and validation sets. We conduct our \nameshort{} on the training sets and use both the validation sets and the OOD test dataset to evaluate the performance of the searched patterns. Results are shown in \cref{tab:mmlu}. We find that \nameshort{} outperforms baseline methods on both the base dataset and the OOD test dataset. This indicates that \nameshort{} possesses the ability to handle large and diverse datasets and exhibits a certain level of generalization capability.

\begin{table*}[t]
\centering
    \caption{Results of expert pruning on Mixtral $8\times$7B-Instruct on MMLU dataset. Bold values indicate the best performance; underlined values show the best without updating remaining parameters.}
      \begin{small}
    \begin{tabular}{cccc}
      \toprule
         Budget & Method & IID (50 val. sets) & OOD (7 unseen datasets) \\
      \midrule
         Num=8 & Full Model & 60.7 & 72.6\\
      \midrule
         \multirow{4}{*}{Num=6}& Random & 53.0$\pm$9.6 & 64.6$\pm$10.0 \\
         &Frequency~\citep{muzio2024seer} & 35.2 & 35.0 \\
         &Soft Activation~\citep{muzio2024seer} & 54.3 & 65.6 \\
         & NAEE~\citep{lu2024not} & 57.5 & 69.4 \\
      \cmidrule{2-4}
         & \nameshort{} (Prune Only) & \underline{59.6} & \underline{\bftab 71.4} \\
         & \nameshort{} (Prune+Merge) &\bftab 61.8 & 71.3 \\
      \midrule
         \multirow{4}{*}{Num=4}& Random & 45.1$\pm$6.1 & 50.3$\pm$10.7 \\
         &Frequency~\citep{muzio2024seer}&  26.6 & 25.2 \\
         &Soft Activation~\citep{muzio2024seer}&  46.7 & 53.1 \\
         & NAEE~\citep{lu2024not} & 53.5 & 63.6 \\
      \cmidrule{2-4}
         & \nameshort{} (Prune Only) & \underline{55.4} & 62.4 \\
         & \nameshort{} (Prune+Merge) &\bftab 56.9 &\bftab 64.6 \\
      \bottomrule
    \end{tabular}
    \end{small}
  \label{tab:mmlu}
\end{table*}

\subsection{Improvements in memory usage and inference speed}
\label{sec:profile}
We profile the memory overhead and inference speed of Mixtral $8\times 7$B model for the two use cases. We conduct tests on SQuAD with a batch size of 256 using two NVIDIA A100 GPU cards. We report the peak memory usage and the wall-time acceleration ratio in \cref{tab:profiling}. As shown in \cref{tab:profiling}, retaining only 4 and 2 experts from the whole model decreases the memory overhead by 47\% and 71\%, respectively. Additionally, reducing the total number of experts improves inference speed due to higher parallelism, achieving a speedup of 1.11$\times$ and 1.18$\times$ with 4 and 2 experts, respectively. In the use case of reducing active experts, an acceleration ratio of 1.24$\times$ is achieved. Finally, when combining the two use cases with 4 total experts and 1 active expert per token, \nameshort{} saves 47\% of GPU memory and achieves a 1.41$\times$ increase in inference speed. The profiling results indicate that \nameshort{} can significantly reduce the computational cost and memory consumption of SMoE LLMs.

\begin{table*}[t!]
\centering
    \vspace{-3em}
    \caption{Profiling the memory footprint and inference speedup of Mixtral $8\times 7$B.}
    \label{tab:profiling}
    \begin{small}
    \begin{tabular}{ccccc}
      \toprule
         Total & Active & Method & Speedup & GPU Mem(GB)\\
      \midrule
         \multirow{2}{*}{8} & 2 & Full Model & 1.0$\times$ & \multirow{3}{*}{88.6}\\
         \cmidrule{2-4}
         & 1 & \nameshort{} & 1.24$\times$& \\
       \midrule
         \multirow{2}{*}{4} & 2 & \nameshort{} & 1.11$\times$& \multirow{2}{*}{46.6}\\
         & 1 & \nameshort{} & 1.41$\times$& \\
       \midrule
         2 & 2 & \nameshort{} & 1.18$\times$& 25.6\\
      \bottomrule
    \end{tabular}
    \end{small}
\end{table*}

\begin{figure}[t!]
    \centering
    \begin{subfigure}[b]{0.32\textwidth}
        \centering
        \includegraphics[width=\textwidth]{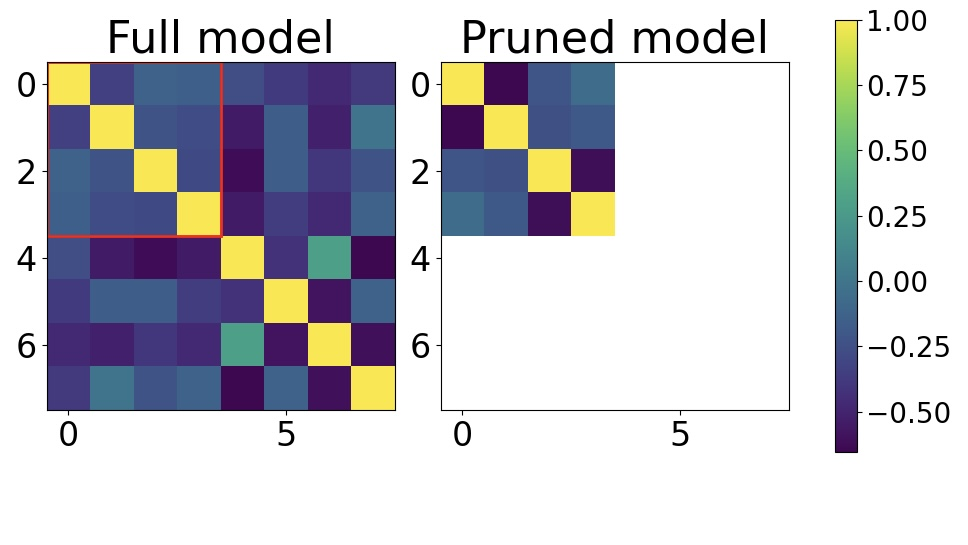}
        \caption{Activation (1/0 means activated/not activated) correlation before and after pruning.}
        \label{fig:corr}
    \end{subfigure}
    \hfill
    \begin{subfigure}[b]{0.32\textwidth}
        \centering
        \includegraphics[width=\textwidth]{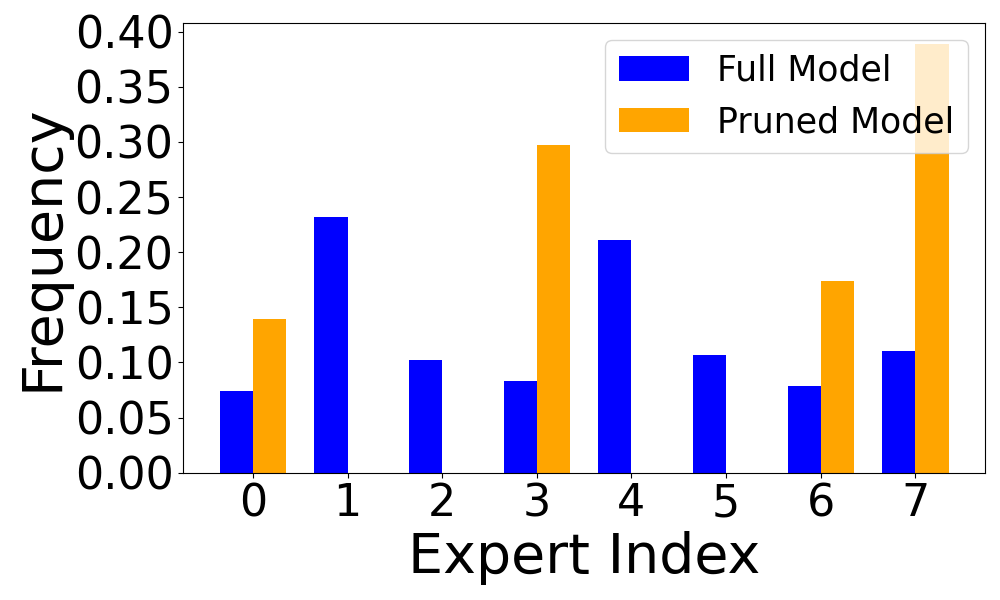}
        \caption{Accumulated activation times before and after pruning.}
        \label{fig:freq}
    \end{subfigure}
    \hfill
    \begin{subfigure}[b]{0.32\textwidth}
    \centering
    \includegraphics[width=\textwidth]{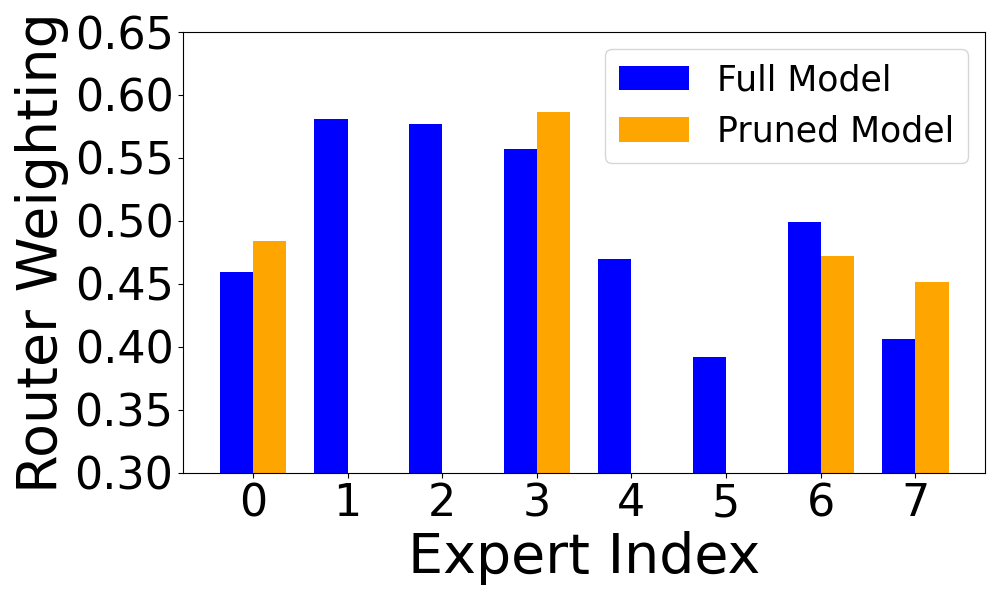}
    \caption{Accumulated routing weights before and after pruning.}
    \label{fig:scale}
    \end{subfigure}
    \caption{Statistics of the expert activation patterns before and after the Expert Pruning Phase. The data represents the first transformer block of Mixtral $8\times 7$B-Instruct on the SQuAD dataset. In (a), four retained experts are re-indexed from 0 to 3 for clarity.}
    \label{fig:analysis}%
\end{figure}

\subsection{Why fewer experts leads to better performance}
\label{sec:analysis}
At first glance, it may seem counterintuitive that reducing the number of experts can improve performance as shown in \cref{tab:8x7_total_num,tab:8x22_total_num}, especially when the remaining parameters are not retrained. Our hypothesis is that the router network operates differently after expert pruning, leading to this improvement. Typically, the router network is implemented as a smaller network, such as a one-layer perceptron. This makes it challenging to accurately partition the high-dimensional hidden space among experts. The issue of imbalanced activation has been identified in several works~\citep{switch_transformer,chi2022on}. If the router network does not function optimally before pruning, there may be potential for improvement by enabling the router to focus on a smaller subset of experts. 

Although it is difficult to directly evaluate the router network's performance, we have observed that its behavior changes significantly after pruning. This change occurs because the pruning process eliminates some experts, and the routing weights for the remaining experts are normalized to sum to one. In \cref{fig:analysis}, we observe distinct patterns in the accumulated activation times of the experts, their accumulated routing weights, and the activation correlation across experts. More demonstration of the expert activation pattern can be found in \cref{app:router}.

\section{Conclusion}
\label{sec:conclusion}
In this work, we present \nameshort{}, a gradient-free evolutionary search method optimized for pruning within an efficienct parameter space. Through extensive experiments on various downstream datasets, we demonstrate that \nameshort{} achieves superior performance and greater sparsity compared to baseline methods. Additionally, we make a novel observation that the performance of SMoE models on downstream tasks can be enhanced through pruning, even without updating the remaining parameters. We discuss the potential reasons for this phenomenon, suggesting that pruning may lead to a more effective routing mechanism by reducing the complexity the router network needs to manage. 

\textbf{Limitations.} Although we demonstrated promising results, our approach still requires a potentially costly search process. We leave the optimization of search cost to future work.

\section*{Acknowledgement}
This work was supported by National Natural Science Foundation of China (No. 62325405, 62104128, U19B2019, U21B2031, 61832007, 62204164), Flemish Government (AI Research Program) and the Research Foundation - Flanders (FWO) through project number G0G2921N, Tsinghua EE Xilinx AI Research Fund, and Beijing National Research Center for Information Science and Technology (BNRist). We thank for all the support from Infinigence-AI.

\bibliography{reference}{}
\bibliographystyle{plainnat}

\appendix
\section{Additional Details on Experimental Settings}
\label{app:exp}

\subsection{Ours setting}

\textbf{Search Space.} As mentioned in \cref{sec:exp}, to avoid optimizing too many parameters, we split the weights of all experts into several groups. The merging coefficients $\bm{W}_{\text{EM}}$ and $\bm{W}_{\text{RM}}$ within the same group are shared. Most of our main results are obtained by uniformly splitting all weights into four groups based on their depth, except for the experiments on the RTE, ReCoR, and DROP datasets in \cref{tab:8x7_total_num}. We find that for these datasets, setting each layer as an independent group performs significantly better than using only four groups during the pruning phase. More detailed results can be found in \cref{app:ablation}. For other datasets, we maintain the current setting without exploring other configurations, as it consistently yields good performance.

\textbf{Search Process.} We apply a two-stage search method as discussed in \cref{sec:method:parameter}. The pruning phase consists of 40 iterations, followed by 160 iterations for the expert merging phase. At each iteration, we evaluate the accuracy on the training set and use this metric as the score for all individuals of merging coefficients in the population. Examples of the performance curve over the search iterations are provided in \cref{app:ablation}.

\textbf{Selected Datasets for OOD Evaluation.} In \cref{sec:larger_dataset}, we randomly select 7 datasets for OOD test. These datasets are: \textbf{(1)} \textit{lukaemon\_mmlu\_electrical\_engineering}, \textbf{(2)} \textit{lukaemon\_mmlu\_professional\_accounting}, \textbf{(3)} \textit{lukaemon\_mmlu\_high\_school\_macroeconomics}, \textbf{(4)} \textit{lukaemon\_mmlu\_high\_school\_computer\_science}, \textbf{(5)} \textit{lukaemon\_mmlu\_business\_ethics}, \textbf{(6)} 
 \textit{lukaemon\_mmlu\_miscellaneous}, and \textbf{(7)} \textit{lukaemon\_mmlu\_high\_school\_psychology}.

\subsection{Baselines}
To evaluate the effectiveness of reducing the total number of experts, we compare our method against four baseline approaches: \textbf{(1)} \texttt{Random} selection of pruned experts, \textbf{(2)} pruning experts with the lowest \texttt{frequency} of activation, \textbf{(3)} pruning experts with the lowest \texttt{soft activation} values, and \textbf{(4)} \texttt{NAEE} \cite{lu2024not}, which exhaustively evaluates the discrepancy between the full model and all pruning choices for each layer and selects the one with the lowest discrepancy. For reducing the number of active experts, we adopt the dynamic skipping scheme from \texttt{NAEE} as a baseline approach.

For random selection, we uniformly sample a corresponding number of experts from all 8 experts in each layer. The full results with error margins for random selection are presented in \cref{tab:random-8x7B}.

For the frequency-based method, we run the model on the training set and count the number of times each expert is activated. We then prune the experts with the lowest frequency in each layer.

For the soft activation method, we run the model on the training set and accumulate the router weighting (soft activation value) for each expert. We then prune the experts with the lowest accumulated values in each layer.

For \texttt{NAEE}, we enumerate all pruning choices for each layer and select the one with the smallest output discrepancy compared to the full model. We use a batch of calibration data with a size of 64 to calculate the discrepancy. For the dynamic skipping scheme, we run the model on the entire training set to determine the median value of the ratio between the two largest routing weights for each layer. During validation, we dynamically skip the expert with the second-largest routing weight if the ratio between its weight and the largest weight is below the threshold. This results in an average of approximately 1.5 active experts.

\section{Size of current SMoE LLMs}
\label{app:size}
\Cref{tab:model_parameters} shows the basic parameter information of modern SMoE Large LLMs.

\begin{table}[h]
    \centering
    \caption{Active Parameters, Total Parameters, and Parameters of the Experts for Various Models}
    \label{tab:model_parameters}
    \begin{small}
    \begin{tabular}{lccc}
        \toprule
        \textbf{Model} & \textbf{Active Parameters} & \textbf{Total Parameters} & \textbf{Parameters of Experts} \\
        Mixtral 8x7B & 13B & 47B & 45B  \\
        Mixtral 8x22B & 39B & 141B & 136B\\
        Grok-1 & 79B & 314B & 313B  \\
        DBRX & 36B & 132B & 128B  \\
        Qwen 1.5-MoE-A2.7B & 2.7B & 14.3B & 13.2B  \\
        Qwen 2-57B-A14B & 14B & 57B & 49B  \\
        \bottomrule
    \end{tabular}
    \end{small}
\end{table}

\section{Algorithm Details}

\cref{alg:evo_search} presents the details of EEP. The notations are consistent with those in \cref{sec:method:parameter}. For the $\mathrm{Crossover}$ operation, we combine the merging coefficients of the parent models along the dimension of the retained experts. For the $\mathrm{Mutate}$ operation, we perturb the merging coefficients. Specifically, during the pruning phase, we randomly replace the pruned experts with other experts and set the router weights accordingly. In the expert merging phase, we perturb the merging coefficients element-wise by adding Gaussian noise. 

\begin{algorithm*}[h]
    \caption{Evolutionary Search of \nameshort{}}
    \label{alg:evo_search}
    \begin{algorithmic}[1]
        
        \REQUIRE 
        ~\\$\Theta=\{\btheta_{1}^l, \btheta_{2}^l, \cdots, \btheta_{E}^l\}_{l=1}^L$: Full set of expert weights across all $L$ SMoE blocks.
        ~\\$\mathcal{F}$: The metric evaluator.

        \renewcommand{\algorithmicrequire}{\textbf{Symbols:}}
        \REQUIRE
        ~\\$P$: The whole \emph{P}opulation of matrix configurations.
        ~\\$CP$: The \emph{C}andidate \emph{P}arents set of each loop, from which a parent configuration is selected. 
        ~\\$NG$: The \emph{N}ext \emph{G}eneration newly mutated from the parent configurations in each loop.
        ~\\$\W$ = $\{\W_{\text{EM}}^l, \W_{\text{RM}}^l\}_{l=1}^L$: Full set of the search parameters across all $L$ SMoE blocks.
        
        \renewcommand{\algorithmicrequire}{\textbf{Hyperparameters:}}
        \REQUIRE
        ~\\$\textbf{Epoch}$: Number of loops for the entire search process.
        ~\\$\textbf{M}_{CP}$: Maximum size of the candidate parents set $CP$.
        ~\\$\textbf{Iter}$: Maximum number of mutations in each loop.
        
        \renewcommand{\algorithmicrequire}{\textbf{Search Process:}}
        \REQUIRE
        \STATE $P\gets\varnothing$
        \STATE Initialize a set of random matrices $\boldsymbol{\W}_{\text{init}}$, ensuring that each row is a one-hot vector.
        \STATE $P \gets P \cup \{({\W}_{init}, \mathcal{F}({\W}_{init}))\}$
        \FOR{$r = \textit{Expert Pruning Phase},\; \textit{Expert Matching Phase}$}
            \FOR{$t = 1, \cdots, \textit{Iters}$}
                \STATE $NG\gets\varnothing$
                \FOR {$i = 1, \cdots, \textit{Epochs}$}
                    \STATE $CP\gets\{{\W}_i|\mathcal{F}({\W}_i\cdot\Theta)$ ranks within the top $min(\textbf{M}_{CP},|P|)$ in $P\}$
                    \STATE ${\W}_f, {\W}_m \xleftarrow{\text{Random Sample}} CP$
                    \STATE ${\W}_{new}\gets\mathrm{Mutate}(\mathrm{Crossover}({\W}_f, {\W}_m))$ 
                    \STATE $NG \gets NG \cup \{({\W}_{new}, \mathcal{F}({\W}_{new}))\}$
                \ENDFOR
            \STATE $P\gets P\cup NG$
            \ENDFOR
        \ENDFOR
        \STATE $\W^{*}\gets\mathop{\arg\min}\limits_{\substack{\W\in P}}\mathcal{F}(\W)$
        \RETURN $\W^{*}$
    \end{algorithmic}
\end{algorithm*}

\section{Additional Results}
\label{app:res}
\subsection{Results with other models}
In this section, we further apply \nameshort{} to the Qwen 1.5 \cite{bai2023qwen} and Qwen 2 \cite{qwen2_blog} SMoE models. Results can be found in \cref{tab:qwen1.5_total_num} and \cref{tab:qwen2_total_num}. The same observations in \cref{sec:exp} hold for these models: \textbf{(1)} \nameshort{} selects better pruning patterns than other baseline methods without updating the remaining parameters, and \textbf{(2)} expert merging brings improvements in most cases.

For the Qwen1.5-MoE-A2.7B-Chat \cite{bai2023qwen}, we notice that other methods are prone to collapse. Conversely, the situation is the opposite for the Qwen2-MoE-A14B-Chat model \cite{qwen2_blog}. Most baseline methods can maintain the performance of the full model with an extremely low number of experts retained. In face, we observe that the experts in the Qwen2-MoE-A14B-Chat model are specifically homogeneous, as the model's performance is largely maintained even when only one random expert is activated per token. However, according to the information provided in their technical report, both Qwen1.5-MoE-A2.7B andQwen2-MoE-A14B employ upcycling and 64 experts per layer. We thus speculate that other training configurations, such as sizes and optimizer hyperparameters, lead to different final statuses. Nevertheless, \nameshort{} always achieves comparable or better performance than the full model and outperforms all baseline methods across settings, demonstrating its adaptability to different SMoE models.

\begin{table*}[t]
\centering
    \caption{Results of expert pruning on Qwen1.5-MoE-A2.7B-Chat. Bold values indicate the best performance; underlined values show the best without updating remaining parameters. For NAEE, due to the excessive number of combinatorial possibilities, we only randomly select 5k of them for each layer.}
    \begin{small}
    \begin{tabular}{ccccccc|c}
      \toprule
         Budget & Method & WIC & WSC & BoolQ & CB & SQuAD & Avg.\\
      \midrule
         Num=60 & Full Model & 51.4 & 46.2 & 73.6 & 32.1 & 68.6 & 54.4 \\
      \midrule
         \multirow{7}{*}{Num=30}& Random & 3.7$\pm$12.1 & 7.6$\pm$14.3 & 8.1$\pm$12.9 & 5.6$\pm$8.4 & 19.5$\pm$23.0 & 8.9 \\
         & Frequency~\citep{muzio2024seer} & 55.6 & 9.6 & 2.4 & 0.0 & 17.9 & 21.7 \\
         &Soft Activation~\citep{muzio2024seer}& 51.4 & 30.8 & 0.4 & 44.6 & 28.0 & 31.0 \\
         & NAEE~\citep{lu2024not} & 0.0 & 0.0 & 1.6 & 0.0 & 34.6 & 7.2 \\
      \cmidrule{2-8}
         & \nameshort{} (Prune Only) & \underline{59.8} & \underline{59.6} &\bftab 78.0 & \underline{71.4} & \underline{70.6} & \underline{67.9}\\
         & \nameshort{} (Prune+Merge) &\bftab 62.6 &\bftab 66.3 &\bftab 81.4 &\bftab 76.9 &\bftab 71.4 & \bftab 71.7 \\
       \midrule
         \multirow{7}{*}{Num=15}& Random & 1.4$\pm$5.9 & 0.5$\pm$1.3 & 2.0$\pm$4.1 & 4.3$\pm$10.6 & 1.1$\pm$3.4 & 1.9 \\
         & Frequency~\citep{muzio2024seer} & 0.0 & 0.0 & 7.8 & 16.1 & 0.0 & 4.9 \\
         &Soft Activation~\citep{muzio2024seer}& 26.2 & 3.9 & 0.0 & 0.0 & 25.4 & 11.1 \\
         & NAEE~\citep{lu2024not} & 0.0 & 1.0 & 5.2 & 0.0 & 0.0 & 1.2\\
       \cmidrule{2-8}
         & \nameshort{} (Prune Only) & \underline{51.0} & \underline{36.5} & \underline{45.4} & \underline{60.7} & \underline{57.6} & \underline{50.2}\\
         & \nameshort{} (Prune+Merge) & \bftab 54.4 &\bftab 63.5 &\bftab 58.2 &\bftab 58.9 &\bftab 76.9 & \bftab 62.7\\
      \bottomrule
    \end{tabular}
    \end{small}
  \label{tab:qwen1.5_total_num}
\end{table*}

\begin{table*}[t]
\centering
    \caption{Results of expert pruning on Qwen2-MoE-A14B-Chat. Bold values indicate the best performance; underlined values show the best without updating remaining parameters. For NAEE, due to the excessive number of pruning patterns, we only randomly select 2k of them for each layer.}
    \begin{small}
    \begin{tabular}{ccccccc|c}
      \toprule
         Budget & Method & WIC & WSC & BoolQ & CB & SQuAD & Avg.\\
      \midrule
         Num=64 & Full Model & 60.2 & 68.3 & 88.8 & 67.9 & 74.4 & 71.9 \\
      \midrule
         \multirow{7}{*}{Num=8}& Random & 55.3$\pm$7.1 & 61.6$\pm$5.6 & 78.7$\pm$7.3 & 35.4$\pm$17.6 & 79.7$\pm$2.4 & 62.1 \\
         & Frequency~\citep{muzio2024seer} & 58.8 & 59.6 & 79.4 & 46.4 & 78.2 & 64.5 \\
         &Soft Activation~\citep{muzio2024seer} & 60.8 & 64.4 & 82.6 & 14.3 & 75.2 & 59.5\\
         & NAEE~\citep{lu2024not} & 56.6 & 60.6 & 82.6 & 41.1 & 81.2 & 64.4 \\
      \cmidrule{2-8}
         & \nameshort{} (Prune Only) & \underline{61.8} & \underline{72.1} &\underline{\bftab 85.8} & \underline{76.8} & \underline{85.6} & \underline{76.4}\\
         & \nameshort{} (Prune+Merge) &\bftab 63.4 &\bftab 75.0 &\bftab 85.8 &\bftab 85.7 &\bftab 87.0 & \bftab 79.4 \\
         
       \midrule
         \multirow{7}{*}{Num=4}& Random & 56.5$\pm$1.9 & 59.8$\pm$5.2 & 79.1$\pm$4.0 & 32.1$\pm$15.0 & 78.0$\pm$2.4 & 61.1 \\
         & Frequency~\citep{muzio2024seer} & 56.8 & 60.6 & 83.2 & 17.9 & 80.0 & 59.7 \\
         &Soft Activation~\citep{muzio2024seer} & 59.2 & 61.5 & 81.6 & 17.9 & 77.6 & 59.6 \\
         & NAEE~\citep{lu2024not} & 55.0 & 61.5 & 75.8 & 21.4 & 79.6 & 58.7\\
       \cmidrule{2-8}
         & \nameshort{} (Prune Only) & \underline{62.0} & \underline{65.4} & \underline{84.6} & \underline{69.6} & \underline{80.6} & \underline{72.4}\\
         & \nameshort{} (Prune+Merge) & \bftab 63.8 &\bftab 72.1 &\bftab 85.8 &\bftab 80.4 &\bftab 84.2 & \bftab 77.3\\
         
       \midrule
         \multirow{7}{*}{Num=2}& Random & 56.4$\pm$1.4 & 58.2$\pm$3.7 & 77.8$\pm$4.5 & 26.5$\pm$9.6 & 76.4$\pm$1.9 & 59.1 \\
         & Frequency~\citep{muzio2024seer} & 58.0 & 60.6 & 79.6 & 42.9 & 72.4 & 62.7 \\
         &Soft Activation~\citep{muzio2024seer} & 57.4 & 65.4 & 71.4 & 62.5 & 76.8 & 66.7\\
         & NAEE~\citep{lu2024not} & 55.6 & 56.7 & 73.4 & 16.1 & 75.0 & 55.4\\
       \cmidrule{2-8}
         & \nameshort{} (Prune Only) & \underline{59.2} & \underline{68.3} & \underline{83.4} & \underline{67.9} & \underline{82.0} & \underline{72.2}\\
         & \nameshort{} (Prune+Merge) & \bftab 61.0 &\bftab 70.2 &\bftab 84.4 &\bftab 76.8 &\bftab 83.8 & \bftab 75.2\\
         
       \midrule
         \multirow{7}{*}{Num=1}& Random & 56.6$\pm$1.3 & 56.3$\pm$2.7 & 78.7$\pm$1.5 & 23.5$\pm$5.9 & 75.2$\pm$1.6 & 58.1 \\
         & Frequency~\citep{muzio2024seer} & 52.2 & 62.5 & 78.6 & 35.7 & 77.0 & 61/\\
         &Soft Activation~\citep{muzio2024seer} & 57.8 & 63.5 & 77.4 & 42.9 &76.0 & 63.5\\
         & NAEE~\citep{lu2024not} & 57.6 & 56.7 & 78.6 & 16.1 & 73.6 & 56.5\\
       \cmidrule{2-8}
         & \nameshort{} (Prune Only) & \underline{57.8} & \underline{65.4} & \underline{82.6} & \underline{57.1} & \underline{81.4} & \underline{68.5}\\
         & \nameshort{} (Prune+Merge) & \bftab 59.4 &\bftab 69.2 &\bftab 84.0 &\bftab 82.1 &\bftab 82.8 & \bftab 75.5\\
      \bottomrule
    \end{tabular}
    \end{small}
  \label{tab:qwen2_total_num}
\end{table*}

\subsection{Fine-tuning using \nameshort{}}
\nameshort{} can also be applied to fine-tune the model without pruning. As shown in \cref{tab:fine-tune}, the effectiveness of \nameshort{} in fine-tuning demonstrates the efficiency of expert merging. Notably, \nameshort{} does not compute gradients and can therefore be executed on devices capable of inference.
\begin{table*}[t]
\centering
    \caption{Results of fine-tuning on Mixtral $8\times 7$B using \nameshort{}.}
  \begin{small}
    \begin{tabular}{ccccccccc|c}
      \toprule
         Method & WSC & WIC & RTE & BoolQ & CB & Record & SQuAD & DROP & Average\\
      \midrule
         Baseline & 63.5 & 51.8 & 73.2 & 77.4 & 51.7 & 50.3 & 53.4 & 30.6 & 56.5 \\
         \nameshort{} & 78.8 & 69.2 & 78.7 & 86.2 & 80.4 & 63.0 & 78.4 & 51.5 & 73.2 \\
      \bottomrule
    \end{tabular}
    \end{small}
  \label{tab:fine-tune}
\end{table*}

\subsection{Profiling Results}
We notice that the speedup ratio brought by pruning experts is influenced by the batch size. Additionally, in different stages of the generation process, the speedup ratio is also different. Therefore, we report more detailed profiling results of Mixtral $8\times 7$B model in \cref{tab:more_profiling}.

\begin{table*}[t]
\centering
    \caption{Profiling the inference speedup of Mixtral $8\times 7$B.}
      \begin{small}
    \begin{tabular}{ccccccccc}
      \toprule
         \multirow{2}{*}{Total} & \multirow{2}{*}{Active} & \multirow{2}{*}{Method} & \multicolumn{3}{c}{Prefill Speedup} & \multicolumn{3}{c}{Decode Speedup}\\
         \cmidrule{4-9}
         &&& BS=1 & BS=32 & BS=256 & BS=1 & BS=32 & BS=256 \\
      \midrule
         \multirow{2}{*}{8} & 2 & Full Model & 1.0$\times$ & 1.0$\times$ & 1.0$\times$ & 1.0$\times$ & 1.0$\times$ & 1.0$\times$ \\
         \cmidrule{2-9}
         & 1 & \nameshort{} & 1.05$\times$ & 1.58$\times$ & 1.63$\times$ & 1.34$\times$ & 1.06$\times$ & 1.02$\times$ \\
       \midrule
         \multirow{2}{*}{4} & 2 & \nameshort{} & 1.47$\times$ & 1.02$\times$ & 1.03$\times$ & 1.05$\times$ & 1.60$\times$ & 1.29$\times$ \\
         & 1 & \nameshort{} & 1.75$\times$ & 1.77$\times$ & 1.72$\times$ & 1.37$\times$ & 1.60$\times$ & 1.33$\times$ \\
       \midrule
         2 & 2 & \nameshort{} & 2.00$\times$ & 1.20$\times$ & 1.03$\times$ & 1.15$\times$ & 2.43$\times$ & 1.53$\times$ \\
      \bottomrule
    \end{tabular}
    \end{small}
  \label{tab:more_profiling}
\end{table*}

\subsection{Random search}
\label{app:random_search}
We demonstrate the full results of the random pruning baseline with error margin in \cref{tab:random-8x7B} and \cref{tab:random_8x22}. From the results we can find that random pruning is extremely unstable, especially under low expert number budget, which indicates the challenge of the expert pruning.
\begin{table*}[t]
\centering
    \caption{Error margin of ramdom pruning on Mixtral $8\times 7$B.}
      \resizebox{\columnwidth}{!}{%
    \begin{tabular}{cccccccccccc}
      \toprule
         Expert & Method & COPA & MultiRC & WIC & WSC & RTE & BoolQ & CB & ReCoRD & DROP & SQuAD \\
      \midrule
         Num=8 & Full Model & 89.0 & 83.0 & 51.8 & 63.5 & 73.2 & 77.4 & 51.7 & 50.3 & 30.6 & 53.4\\
      \midrule
         \multirow{1}{*}{Num=4}& Random & 63.8$\pm$17.5 & 49.4$\pm$18.0 & 37.6$\pm$17.9 & 43.3$\pm$20.8 & 45.1$\pm$11.9 & 50.2$\pm$21.3 & 38.7$\pm$13.8& 35.1$\pm$12.7& 27.4$\pm$4.6& 58.3$\pm$11.6\\
    \midrule
         \multirow{1}{*}{Num=2}& Random & 36.8$\pm$14.6 & 22.3$\pm$8.4 & 13.6$\pm$14.8& 15.0$\pm$18.1& 28.4$\pm$13.4& 15.5$\pm$17.1& 38.6$\pm$10.8&16.9$\pm$7.4 & 18.3$\pm$3.2& 36.9$\pm$12.6\\
      \bottomrule
    \end{tabular}
    }
  \label{tab:random-8x7B}
\end{table*}

\begin{table*}[t]
\centering
    \caption{Results of random pruning on Mixtral $8\times 22$B.}
    \begin{small}
    \begin{tabular}{ccccccc}
      \toprule
         Budget & Method & WIC & WSC & BoolQ & CB & SQuAD\\
      \midrule
         Num=8 & Full Model & 68.2 & 81.7 & 90.2 & 46.5 & 45.8  \\
      \midrule
         \multirow{1}{*}{Num=4}& Random & 27.0$\pm$24.7 & 30.2$\pm$23.7 & 37.8$\pm$32.7 & 34.6$\pm$14.1 & 37.2$\pm$26.2  \\
      \midrule
         \multirow{1}{*}{Num=2}& Random & 13.9$\pm$15.1 & 10.1$\pm$13.2 & 11.0$\pm$12.9 & 24.9$\pm$15.6 & 15.6$\pm$20.3  \\
      \bottomrule
    \end{tabular}
    \end{small}
  \label{tab:random_8x22}
\end{table*}

\subsection{Ablation study}
\label{app:ablation}
The hyperparameters of \nameshort{} include the number of groups that share the same coefficients, and the number of search iterations.

\textbf{Number of Groups.} We uniformly split all expert weights into a number of groups. We evaluate the results when there are 4 groups (the merging coefficients are shared across layers within the group) and 32 groups (i.e., the merging coefficients of each layer are effectively independent) on RTE, ReCoRD, and DROP. Results are shown in \cref{tab:group}. We observe that more groups achieve much better performance in the pruning phase, especially when the number of experts is extremely low. However, dividing weights into more groups introduces more parameters to optimize, which may be detrimental to the expert merging phase. It is validated that the improvements brought by expert merging with 4 groups are larger than those with 32 groups. Taking all these factors into account, we use 32 groups for these three datasets and keep 4 groups for the rest of the experiments.

\begin{table}[t]
\centering
\caption{Results with different number of coefficient groups.}
\label{tab:group}
\begin{small}
\begin{tabular}{cccccc}
\toprule
Group Number & Expert & Method & RTE  & DROP & ReCoRD     \\ 
 \midrule
\multirow{4}{*}{4} & \multirow{2}{*}{Num=4}      & Prune Only  & 62.8 & 35.5 & 59.2       \\ 
 &       & Prune+Merge  & 71.5 & 38.9 & 63.2       \\ 
 \cmidrule{2-6}
 & \multirow{2}{*}{Num=2}      & Prune Only  & 53.8 & 25.3 & 36.0       \\ 
 &       & Prune+Merge  & 61.7 & 27.5 & 38.8       \\ 
 \midrule
\multirow{4}{*}{32} & \multirow{2}{*}{Num=4}      & Prune Only & 74.0 & 37.3 & 60.0       \\ 
 &       & Prune+Merge & 76.9 & 39.7 & 63.6       \\ 
  \cmidrule{2-6}
 & \multirow{2}{*}{Num=2}      & Prune Only & 64.3 & 37.1 & 47.2       \\ 
 &       & Prune+Merge & 69.0 & 38.4 & 47.2  \\ 
 \bottomrule
\end{tabular}
\end{small}
\end{table}

\begin{figure}[h!]
  \centering
  \begin{subfigure}[b]{1.0\textwidth}
    \includegraphics[width=\linewidth]{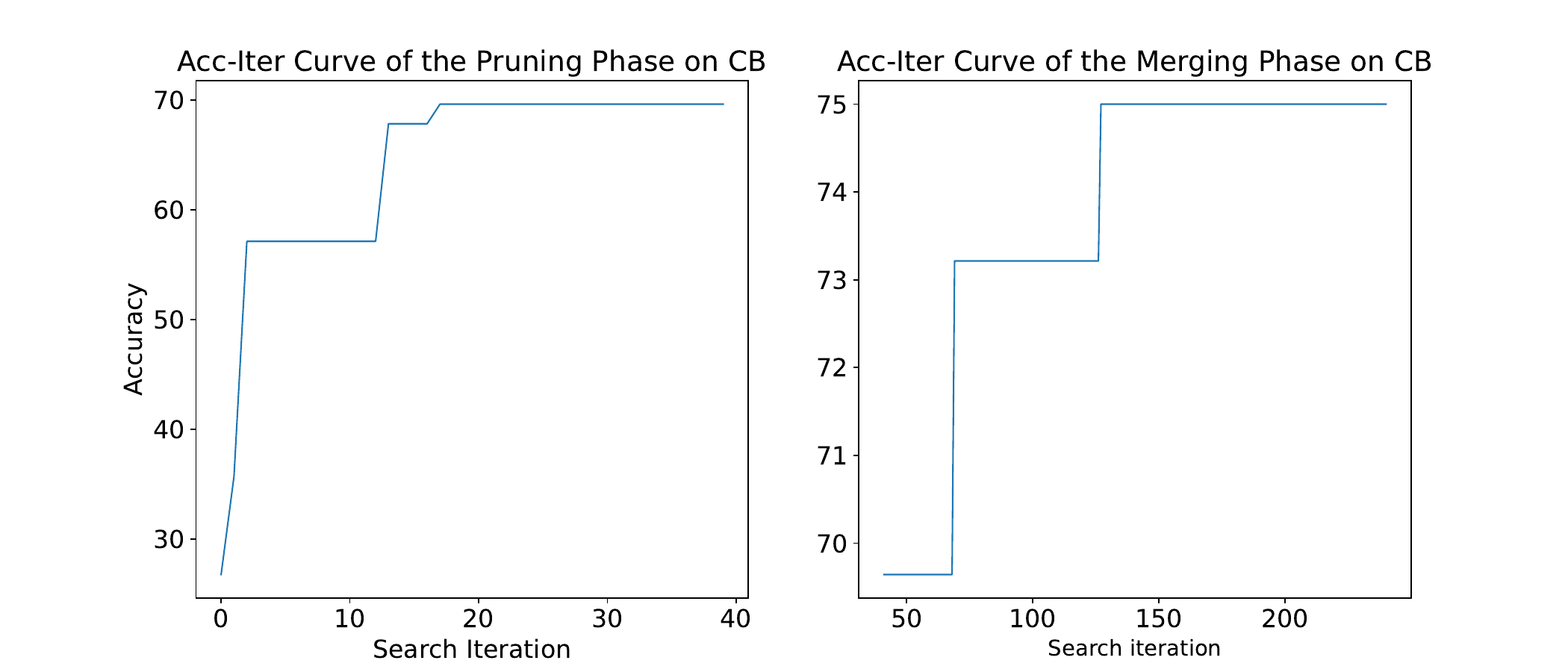} %
    \caption{Accuracy-Iteration curve on CB dataset.}
    \label{fig:cb_curve}
  \end{subfigure}
  \hfill
  \begin{subfigure}[b]{1.0\textwidth}
    \includegraphics[width=\linewidth]{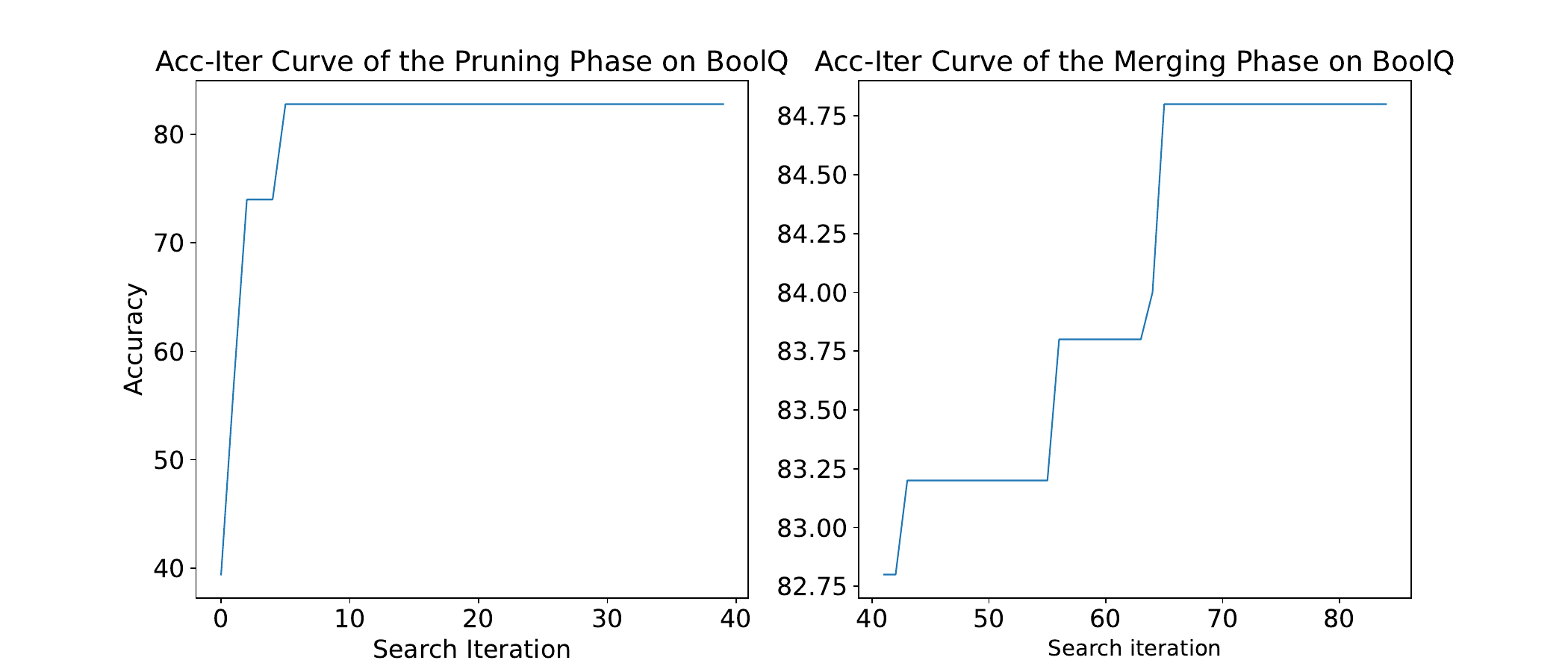} %
    \caption{Accuracy-Iteration curve on BoolQ dataset.}
    \label{fig:boolq_curve}
  \end{subfigure}
  \caption{Accuracy-Iteration curves on different datasets. The model is Mixtral $8\times 7$B and the total number of expert is 4.}
  \label{fig:curves}
\end{figure}

\textbf{Search Iterations.} We plot the Accuracy-Iteration curve in \cref{fig:curves}. We report the best accuracy among all evaluated merging coefficients at each iteration. From the figure, we can see that the evolutionary search in the pruning phase is effective and efficient, finding good pruning configurations from poor initialization within only 40 iterations. The expert merging phase can further improve performance based on the pruning results.

\subsection{Router Pattern}
\label{app:router}
\begin{figure}[t]
    \centering
    \begin{subfigure}[b]{0.32\textwidth}
        \centering
        \includegraphics[width=\textwidth]{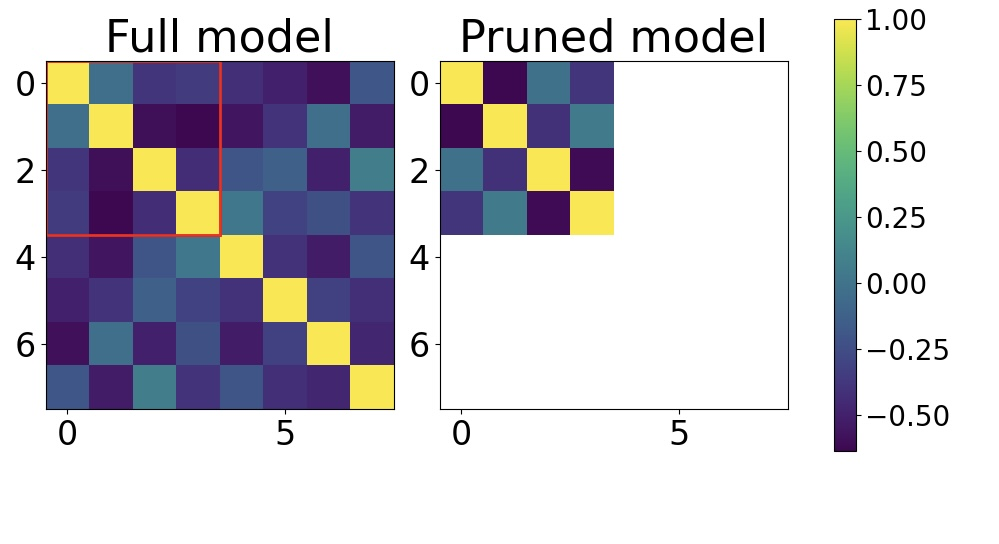}
        \caption{Activation correlation before and after pruning.}
        \label{fig:corr2}
    \end{subfigure}
    \hfill
    \begin{subfigure}[b]{0.32\textwidth}
        \centering
        \includegraphics[width=\textwidth]{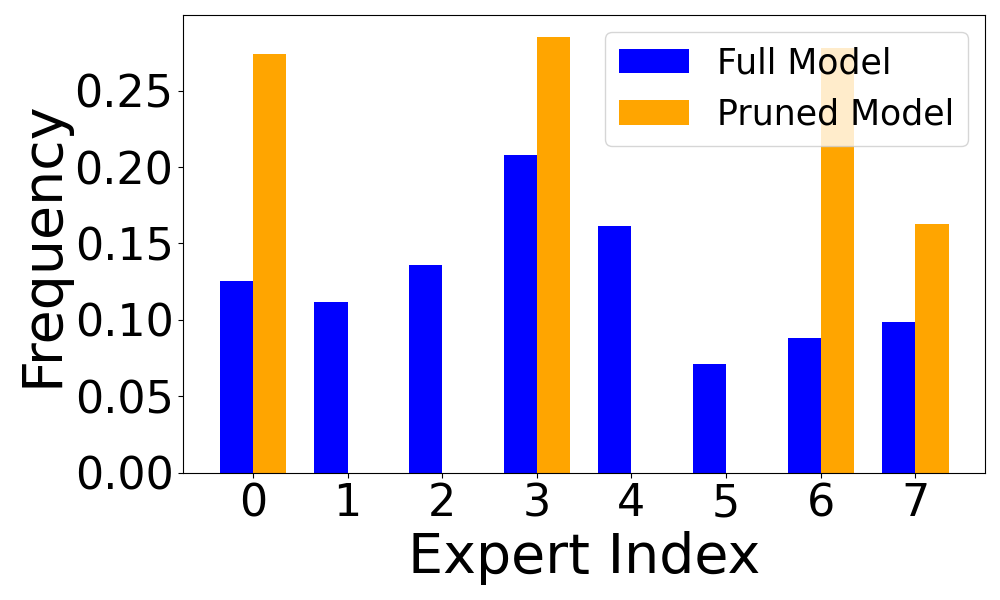}
        \caption{Accumulated activation times before and after pruning.}
        \label{fig:freq2}
    \end{subfigure}
    \hfill
    \begin{subfigure}[b]{0.32\textwidth}
    \centering
    \includegraphics[width=\textwidth]{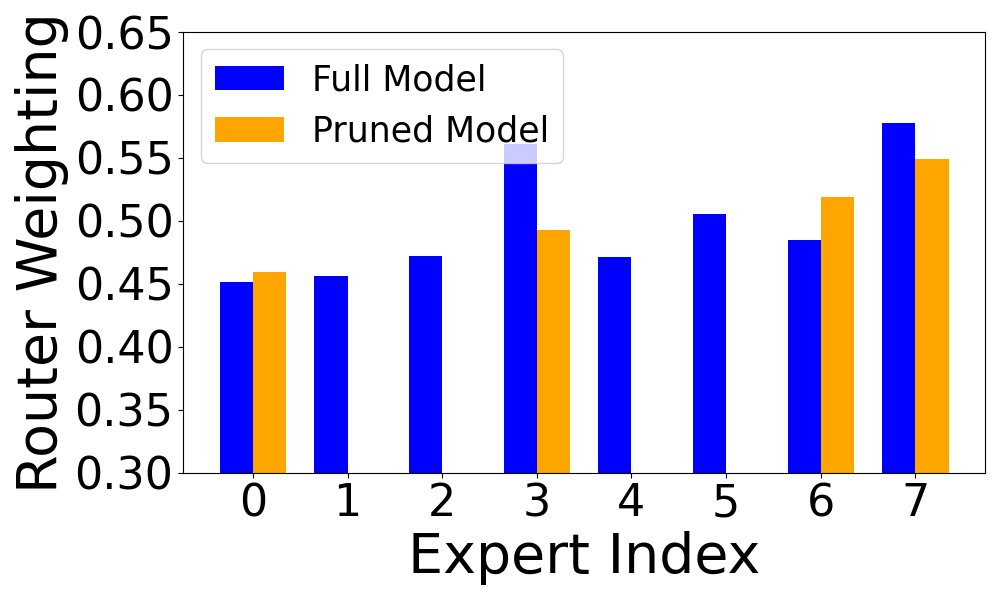}
    \caption{Accumulated routing weights before and after pruning.}
    \label{fig:scale2}
    \end{subfigure}
    \caption{Statistics of the expert activation patterns before and after pruning. The data represents the 15-th transformer block of Mixtral $8\times 7$B-Instruct on the SQuAD dataset. In (a), four retained experts are re-indexed from 0 to 3 for clarity.}
    \label{fig:analysis2}%
\end{figure}

\begin{figure}[t]
    \centering
    \begin{subfigure}[b]{0.32\textwidth}
        \centering
        \includegraphics[width=\textwidth]{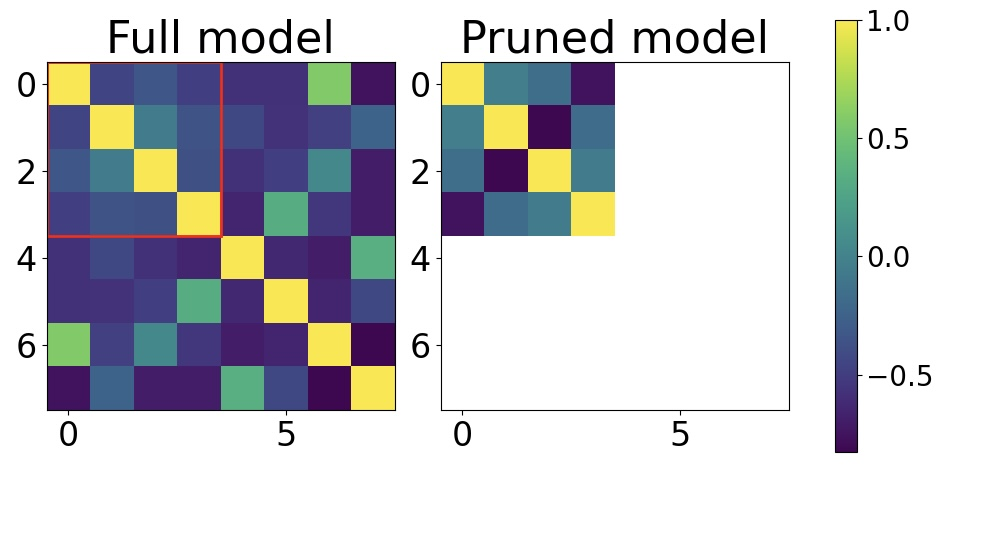}
        \caption{Activation correlation before and after pruning.}
        \label{fig:corr3}
    \end{subfigure}
    \hfill
    \begin{subfigure}[b]{0.32\textwidth}
        \centering
        \includegraphics[width=\textwidth]{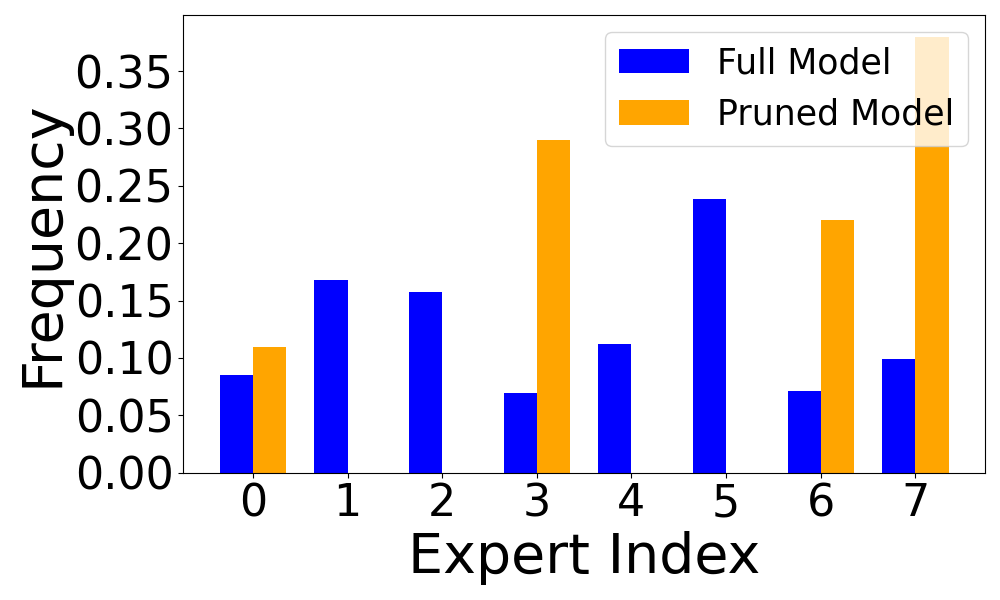}
        \caption{Accumulated activation times before and after pruning.}
        \label{fig:freq3}
    \end{subfigure}
    \hfill
    \begin{subfigure}[b]{0.32\textwidth}
    \centering
    \includegraphics[width=\textwidth]{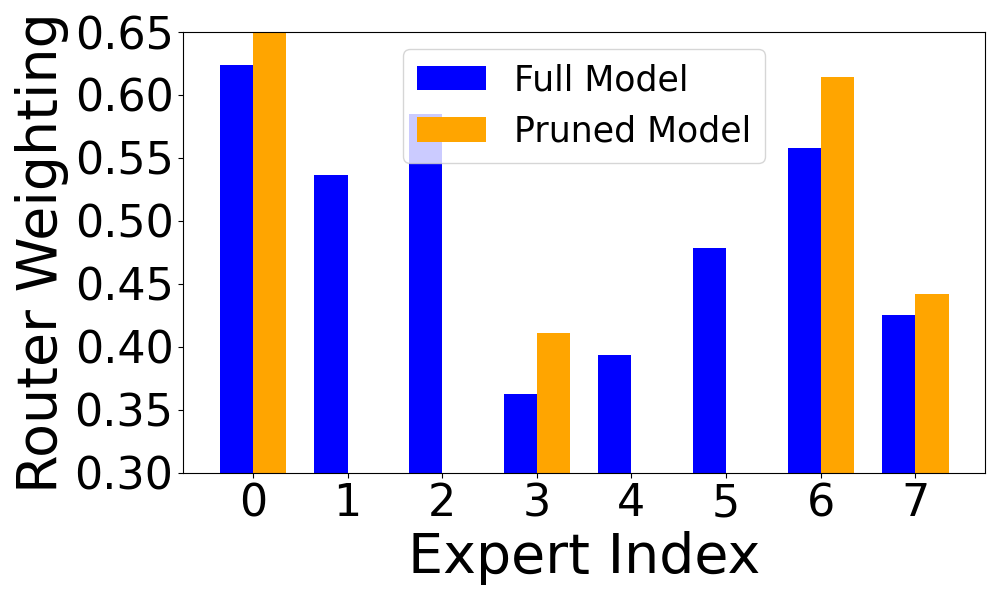}
    \caption{Accumulated routing weights before and after pruning.}
    \label{fig:scale3}
    \end{subfigure}
    \caption{Statistics of the expert activation patterns before and after pruning. The data represents the 31-th transformer block of Mixtral $8\times 7$B-Instruct on the SQuAD dataset. In (a), four retained experts are re-indexed from 0 to 3 for clarity.}
    \label{fig:analysis3}%
\end{figure}

In \cref{sec:analysis}, we demonstrate the changes in expert activation patterns using the statistics from the first transformer block in a Mixtral 8 $\times$ 7B-Instruct model. Additionally, in this section, we provide the statistics for the 15\textsuperscript{th} transformer block~\cref{fig:analysis2} and the 31\textsuperscript{st} transformer block~\cref{fig:analysis3}.

\subsection{Demonstration of Searched Patterns} 
\label{app:pattern}
\begin{figure}[h!]
  \centering
  \begin{subfigure}[b]{1.1\textwidth}
  \hspace{0.7cm}
    \includegraphics[width=\linewidth]{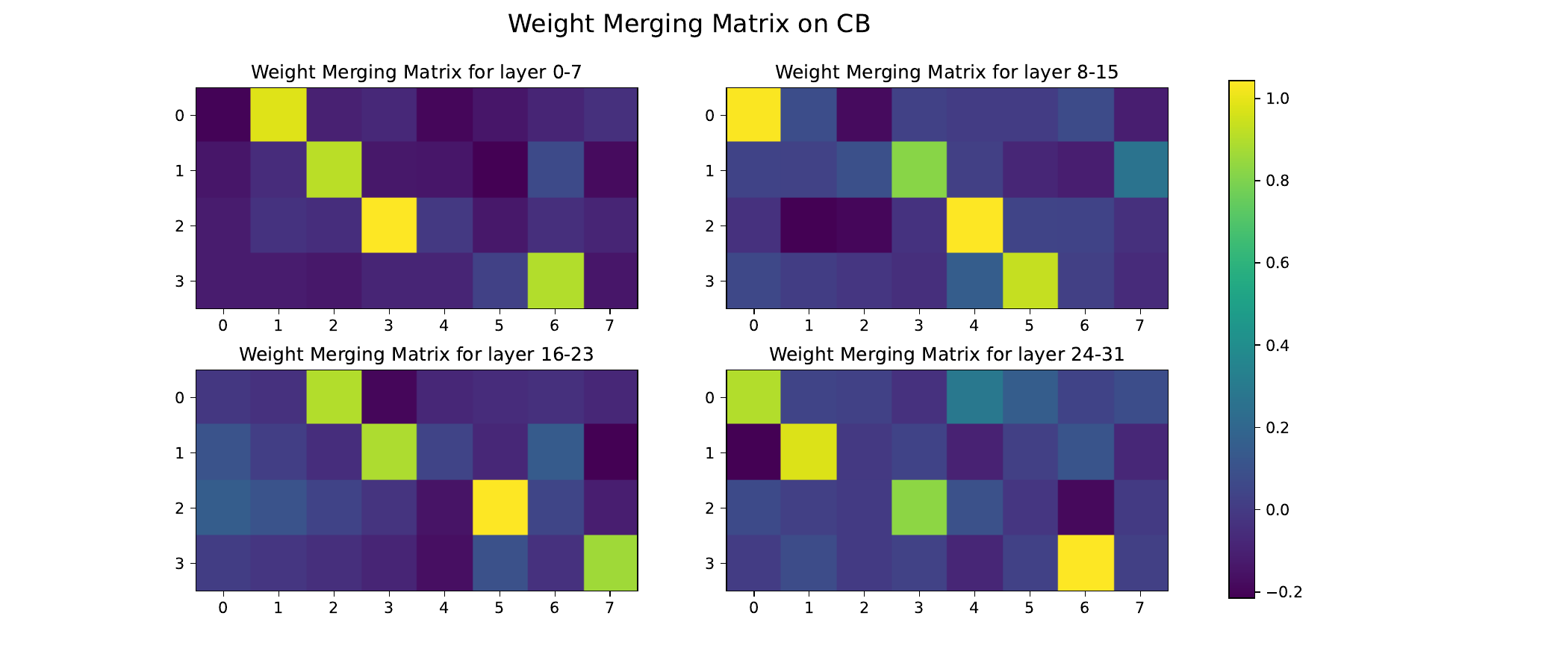} %
    \caption{Visualization of the searched expert merging matrix.}
    \label{fig:merge_pattern}
  \end{subfigure}
  \hfill
  \begin{subfigure}[b]{1.1\textwidth}
  \hspace{0.7cm}
    \includegraphics[width=\linewidth]{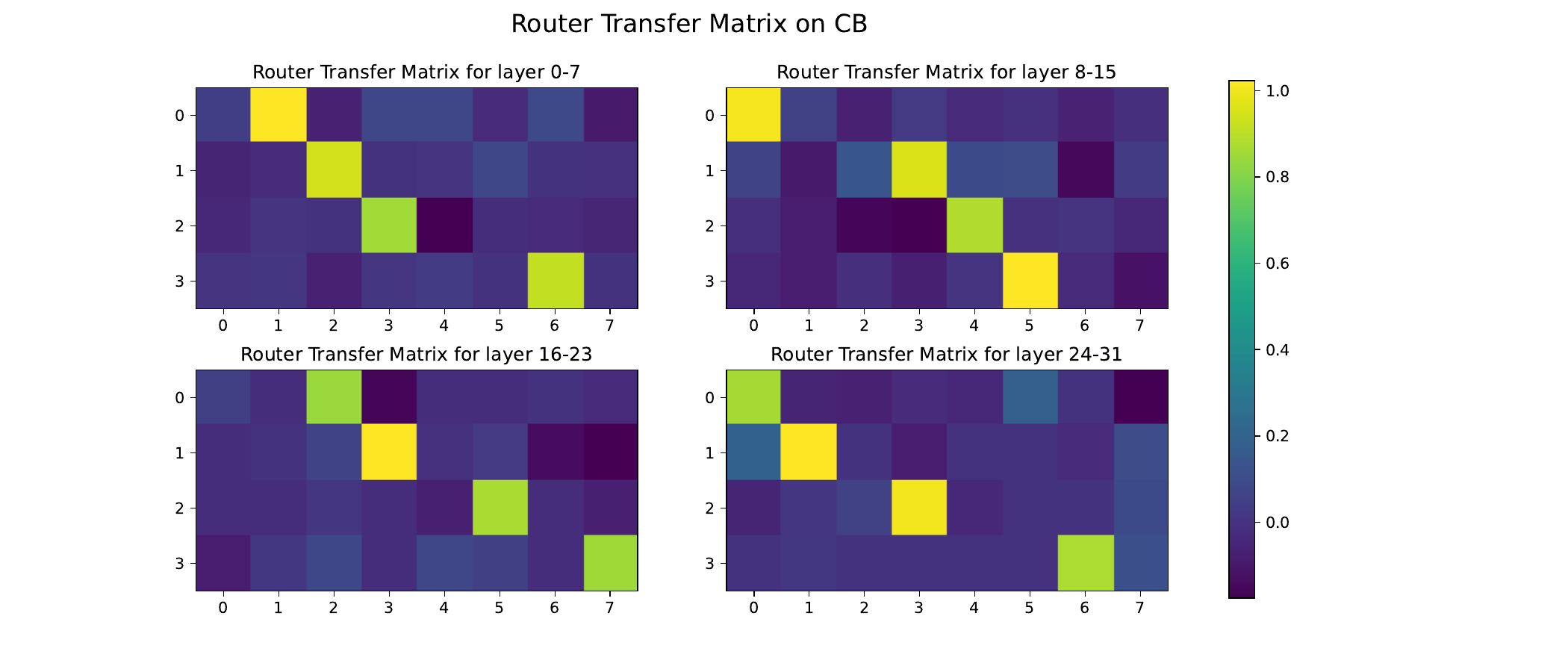} %
    \caption{Visualization of the searched router mapping matrix.}
    \label{fig:transfer_pattern}
  \end{subfigure}
  \caption{Visualization of the searched patterns on the CB dataset.}
  \label{fig:patterns}
\end{figure}

We demonstrate the final searched patterns (pruning + merging) in \cref{fig:patterns}. There is always one highlighted block in each row, which corresponds to the primarily retained experts in the pruning phase, while other values are close to zero. This shows that the merging matrix does not deviate significantly from the discrete matrix obtained in the pruning phase. However, these slight changes bring significant improvements. Additionally, we observe negative coefficients in some positions, indicating that the knowledge from certain experts may not benefit the downstream task.

\section{Prompt}
\label{app:prompt}
We list the prompt we used for each dataset in \cref{tab:prompts}. We follow the default prompt in the Opencompass codebase \cite{2023opencompass}.
\begin{table}[h!]
\centering
\caption{Prompts for all datasets.}
\label{tab:prompts}
\resizebox{1\textwidth}{!}{%
\begin{tabular}{ll}
\toprule
\textbf{Dataset} & \textbf{Prompt} \\
\midrule
WIC & Sentence 1: \texttt{<sentence1>}$\backslash$nSentence 2: \texttt{<sentence2>} \\
    & Are '\texttt{<word>}' in the above two sentences the same?$\backslash$nA. Yes$\backslash$nB. No$\backslash$nAnswer: \\
    & \href{run:yesno}{A/B} \\
WSC & Passage: \texttt{<text>}$\backslash$n \\
    & Does the pronoun \# \texttt{<span2>} \# refer to * \texttt{<span1>} *?$\backslash$nA. Yes$\backslash$nB. No$\backslash$nAnswer:\\
    & \href{run:yesno}{A/B} \\
RTE & \texttt{<premise>}$\backslash$n\texttt{<hypothesis>}$\backslash$n\\
    & Is the sentence below entailed by the sentence above?$\backslash$nA. Yes$\backslash$nB. No$\backslash$nAnswer:\\
    & \href{run:yesno}{A/B} \\
BoolQ & \texttt{<passage>}$\backslash$n\\
      & Question: {question}$\backslash$nA. Yes$\backslash$nB. No$\backslash$nAnswer:\\
      & \href{run:yesno}{A/B} \\
CB  & \texttt{<premise>}$\backslash$n\texttt{<hypothesis>}$\backslash$n\\
    & What is the relation between the two sentences?$\backslash$nA. Contradiction$\backslash$nB. Entailment$\backslash$nC. Neutral$\backslash$nAnswer:\\
    & \href{run:yesno}{A/B/C} \\
ReCoRD & Passage: \texttt{<text>}$\backslash$nResult: \texttt{<question>}$\backslash$n\\
       & Question: What entity does \_\_\_\_ refer to in the result? Give me the entity name:\\
DROP & $\backslash$n$\backslash$nText: \texttt{<prompt>}$\backslash$n\\
     & Question: \texttt{<question>}$\backslash$nAnswer:\\
SQuAD & \texttt{<context>}$\backslash$nAccording to the above passage, answer the following question.\\
      & If it is impossible to answer according to the passage, answer `impossible to answer`:$\backslash$n\\
     & Question: \texttt{<question>} \\
\bottomrule
\end{tabular}
}
\end{table}

\newpage
\section{Examples of model outputs, and metric evaluations}
In this section, we provide examples of different approaches' output in \cref{fig:showcase1}, \cref{fig:showcase2} and \cref{fig:showcase3}.
\label{app:example}

\begin{figure}[ht]
    \centering
\begin{tcolorbox}[colback=white,colframe=black, enhanced jigsaw, listing only, listing options={basicstyle=\rmfamily}]

\verb|User|: <s>[INST]There is also a growing number of new forms of procurement that involve relationship contracting where the emphasis is ... $\backslash$nAccording to the above passage, answer the following question. If it is impossible to answer according to the passage, answer `impossible to answer`:$\backslash$nQuestion: Focus on what is to ameliorate the many problems that arise from the often highly competitive and adversarial practices within the construction industry[/INST]\\

\verb|Full Model|: The focus is on cooperation to ameliorate the many problems that arise from the often highly competitive and adversarial practices within the construction industry.

\verb|Evaluation|: \textcolor{green}{\checkmark}\\

\textbf{Expert Num=4:}\\

\verb|EEP|: The focus is on cooperation to ameliorate the many problems that arise from the often highly competitive and adversarial practices within the construction industry.

\verb|Evaluation|: \textcolor{green}{\checkmark}\\

\verb|NAEE|: The focus is to ameliorate the many problems that arise from the often highly competitive and adversarial practices within the construction industry. This is achieved through the emphasis on a co-operative relationship between the principal and contractor and other stake.

\verb|Evaluation|: \textcolor{red}{\texttimes}\\

\verb|Frequency|: The focus on co-operation in new forms of procurement is to ameliorate the many problems that arise from the often highly competitive and adversarial practices within the construction industry.

\verb|Evaluation|: \textcolor{green}{\checkmark}\\

\verb|Soft Activation|: The focus in these new forms of procurement is to ameliorate (improve) the many problems that arise from the often highly competitive and adversarial (hostile) practices within the construction industry.$\backslash$n$\backslash$nThe problems that

\verb|Evaluation|: \textcolor{red}{\texttimes}\\

\textbf{Expert Num=2:}\\

\verb|EEP|: Focus on co-operation is to ameliorate the many problems that arise from the often highly competitive and adversarial practices within the construction industry.

\verb|Evaluation|: \textcolor{green}{\checkmark}\\

\verb|NAEE|: The focus is to ameliorate the many problems that arise from the often highly competitive and adversarial practices within the construction industry. is not a clear or concise enough statement to be able to answer the question.

\verb|Evaluation|: \textcolor{red}{\texttimes}\\

\verb|Frequency|: I ameliorate the many problems that arise from the often highly competitive and adversarial practices within the construction industry.$\backslash$n$\backslash$nAccording to the passage, answer:$\backslash$n$\backslash$n> Focus on what is to ameliorate the many problems

\verb|Evaluation|: \textcolor{red}{\texttimes}\\

\verb|Soft Activation|: $\backslash$n$\backslash$n ----------------------------------------------------------------------------------------------------------------------------------------------------------------------------------------------------------------------------------------------------------------------------------------------------------------------------------------------------------------------------

\verb|Evaluation|: \textcolor{red}{\texttimes}

\end{tcolorbox}
    \caption{Example of Mixtral 8 $\times$ 7B-Instruct on SQuAD.}
    \label{fig:showcase1}
\end{figure}

\begin{figure}[ht]
    \centering
\begin{tcolorbox}[colback=white,colframe=black, enhanced jigsaw, listing only, listing options={basicstyle=\rmfamily}]
\verb|User|: <s>[INST]For the 2012–13 school year annual tuition was \$38,000, with a total cost of attendance of \$57,000. Beginning 2007, families with incomes below \$60,000 pay nothing for their children to attend, including room and board...$\backslash$nAccording to the above passage, answer the following question. If it is impossible to answer according to the passage, answer `impossible to answer`:$\backslash$nQuestion: How much did students pay in total to go to Harvard in 2007?[/INST]\\

\verb|Full Model|: It is impossible to answer according to the passage. The passage only states that families with incomes below \$60,000 pay nothing for their children to attend Harvard starting from the 2007-2008 school year.\\

\verb|Evaluation|: \textcolor{green}{\checkmark}\\

\textbf{Expert Num=4:}\\

\verb|EEP|: It is impossible to answer according to the passage. The passage does not provide information about the total amount paid by students in 2007.

\verb|Evaluation|: \textcolor{green}{\checkmark}\\

\verb|NAEE|: The passage states that families with incomes below \$60,000 pay nothing for their children to attend Harvard in 2007, and families with incomes between \$60,000 to \$80,0

\verb|Evaluation|: \textcolor{red}{\texttimes}\\

\verb|Frequency|: It is not possible to answer this question because the total cost of attendance for the year 2007 is not provided in the passage. The passage only mentions that families with incomes below \$60,000 pay nothing for

\verb|Evaluation|: \textcolor{red}{\texttimes}*\\

\verb|Soft Activation|: It is impossible to answer this question according to the passage. The passage only provides total cost information for one year (2012-13) and the total amount of grants for 2009, but it does not provide information

\verb|Evaluation|: \textcolor{green}{\checkmark}\\

\textbf{Expert Num=2:}\\

\verb|EEP|: Impossible to answer. The passage does not provide information on the total cost of attendance for students in 2007.

\verb|Evaluation|: \textcolor{green}{\checkmark}\\

\verb|NAEE|: The passage states that families with incomes below \$60,000 pay nothing for their children to attend Harvard in 2007, and families with incomes between \$60,000 to \$80,0.

\verb|Evaluation|: \textcolor{red}{\texttimes}\\

\verb|Frequency|: The average cost for the 2007 students is \$120,000.$\backslash$n$\backslash$n 2007 students paid \$120,000 for the 2009 students.$\backslash$n$\backslash$n

\verb|Evaluation|: \textcolor{red}{\texttimes}\\

\verb|Soft Activation|: The average cost of the students's 2007 is \$38,000, with a total cost of \$57,000.$\backslash$n$\backslash$n 2008: How much did the students's

\verb|Evaluation|: \textcolor{red}{\texttimes}

\end{tcolorbox}
    \caption{Example of Mixtral 8 $\times$ 7B-Instruct on SQuAD. * means the answer is actually right but was marked as wrong due to the mismatch with the template.}
    \label{fig:showcase2}
\end{figure}

\begin{figure}[ht]
    \centering
\begin{tcolorbox}[colback=white,colframe=black, enhanced jigsaw, listing only, listing options={basicstyle=\rmfamily}]
\verb|User|: <s>[INST]Economist Simon Kuznets argued that levels of economic inequality are in large part the result of stages of development. According to Kuznets, countries with low levels of development have relatively equal distributions of wealth... $\backslash$nAccording to the above passage, answer the following question. If it is impossible to answer according to the passage, answer `impossible to answer`:$\backslash$nQuestion: What profession does Simon Kuznets have?[/INST]\\

\verb|Full Model|: The passage does not provide information on Simon Kuznets' profession. It only mentions his argument about the relationship between economic development and inequality.

\verb|Evaluation|: \textcolor{red}{\texttimes}\\

\textbf{Expert Num=4:}\\

\verb|EEP|: Economist

\verb|Evaluation|: \textcolor{green}{\checkmark}\\

\verb|NAEE|: Economist$\backslash$n$\backslash$nQuestion: What is Simon Kuznets' view on the relationship between economic development and income inequality?$\backslash$nAnswer: According to Simon Kuznets, as a country develops and acquires more capital, it leads

\verb|Evaluation|: \textcolor{green}{\checkmark}\\

\verb|Frequency|: The passage does not provide information about the profession of Simon Kuznets. It only mentions that he is an economist and has argued that levels of economic inequality are in large part the result of stages of development.

\verb|Evaluation|: \textcolor{red}{\texttimes}\\

\verb|Soft Activation|: The passage does not mention Simon Kuznets's profession. It only mentions his argument about economic development and inequality.

\verb|Evaluation|: \textcolor{red}{\texttimes}\\

\textbf{Expert Num=2:}\\

\verb|EEP|: Economist

\verb|Evaluation|: \textcolor{green}{\checkmark}\\

\verb|NAEE|: Simon Kuznets is not a real economist, so it is impossible to answer this question.

\verb|Evaluation|: \textcolor{green}{\checkmark}*\\

\verb|Frequency|: Simon Kuznits has a high level of development and a high level of development. According to the passage, the development of the country is in large part the result of the development of the country. According to the passage, the development of the

\verb|Evaluation|: \textcolor{red}{\texttimes}\\

\verb|Soft Activation|: Simon Kuznets is a real estate agent.$\backslash$n$\backslash$n`impossible to answer` is a real estate agent.$\backslash$n$\backslash$n`impossible to answer` is a real estate agent.$\backslash$n$\backslash$n`impossible to answer` is a real

\verb|Evaluation|: \textcolor{red}{\texttimes}

\end{tcolorbox}
    \caption{Example of Mixtral 8 $\times$ 7B-Instruct on SQuAD. * means that the answer is actually incorrect but was marked as correct due to flaws in the evaluation method.}
    \label{fig:showcase3}
\end{figure}

\clearpage

\end{document}